\pdfoutput=1
\documentclass{article}

\usepackage{iclr2027_conference,times}

\usepackage{amsmath,amsfonts,bm}

\def\eqref#1{equation~\ref{#1}}
\def\1{\bm{1}}

\DeclareMathAlphabet{\mathsfit}{\encodingdefault}{\sfdefault}{m}{sl}
\SetMathAlphabet{\mathsfit}{bold}{\encodingdefault}{\sfdefault}{bx}{n}

\usepackage{hyperref}
\usepackage{url}

\usepackage{graphicx}
\usepackage{booktabs}
\usepackage{amsmath}
\usepackage{amssymb}
\usepackage{xcolor}
\usepackage{microtype}
\usepackage{float}
\usepackage{placeins}
\usepackage{tabularx}
\usepackage{xspace}
\usepackage{booktabs}
\usepackage{array}

\newcolumntype{L}[1]{>{\raggedright\arraybackslash}p{#1}}
\newcolumntype{Y}{>{\raggedright\arraybackslash}X}

\newcommand{\rev}[1]{{#1}}

\DeclareRobustCommand{\Dxiii}{\textup{D\kern0.03em\scalebox{0.85}{13}}\xspace}
\title{{Blind Denoiser Selection Without Noise Identification}}

\author{
  Giovanni Nocerino\textsuperscript{1\,$\dagger$},
  Matteo Colantoni\textsuperscript{2\,$\dagger$},
  Rub\'en S\'anchez-Garc\'ia\textsuperscript{3}, and
  Luciano Dyballa\textsuperscript{3}\thanks{Corresponding author: \texttt{luciano.dyballa@ie.edu}} \\[1ex]
  \textsuperscript{1}Bocconi University, Milan, Italy \\
  \textsuperscript{2}London School of Economics and Political Science, London, UK \\
  \textsuperscript{3}School of Science \& Technology, IE University, Madrid, Spain
}

\iclrfinalcopy

\begin{document}

\maketitle

{\renewcommand{\thefootnote}{\fnsymbol{footnote}}\footnotetext[2]{Equal contribution.}}

\begin{abstract}
\rev{Denoisers are increasingly specialized, each strong on the noise it was designed or trained for and weak elsewhere. When many images of unknown origin must be denoised without clean references, as in an automated pipeline, inspecting the output of every candidate is impractical, and the usual choice is to apply the denoiser that performs best on average, which in the pools we study falls 1 to 2.5~dB short of the best choice for each image. Existing methods close this gap by identifying the corruption first or by running every candidate and scoring the outputs. We show that choosing a denoiser does not require identifying the corruption, because corruptions that differ often call for the same denoiser. We label a set of calibration images by the candidate that denoises each one best rather than by its corruption, describe every image by thirteen statistics of its noise, and send a new image to the denoiser of its nearest calibration image. In this space, an image's nearest neighbor has the same best denoiser in 83 to 89\% of cases but the same corruption in only 60\%. On corruptions withheld from calibration, routing recovers 43 to 66\% of the improvement available from a perfect per-image choice, with one denoising call and no retraining, and it outperforms seven selectors that run and score every candidate. Learned routers trained on the same labels do no better, so a simple designed descriptor is sufficient once the calibration images are labeled by their best denoiser. On real camera noise, which lies far from every calibration image, the descriptor flags most images before any denoiser is run, and adding labeled references removes most of the loss without retraining. For JPEG compression, adding references together with one new denoiser brings routing within 0.25~dB of a perfect per-image choice.}
\end{abstract}

\section{Introduction}

Denoisers differ in the noise they are designed for. \rev{A classical filter assumes a noise model, such as a fixed variance \citep{dabov2007image} or impulses at isolated pixels \citep{hwang1995adaptive}, and a trained network handles the degradations it was shown during training \citep{zhang2017beyond}.} A denoiser run on noise outside its intended range either erases detail that belongs to the scene or leaves the corruption in place. \rev{When many images of unknown origin must be denoised without clean references, as in an automated pipeline, inspecting the output of every candidate is impractical, and the usual choice is to apply the denoiser that performs best on average. In the pools of denoisers we study, this falls 1 to 2.5~dB short of the best choice for each image. Existing methods that choose per image either identify the corruption and apply the denoiser matched to it, or run every candidate and score the outputs without a reference.}

\rev{We show that choosing a denoiser does not require identifying the corruption, because corruptions that differ often call for the same denoiser. A notch filter, for example, removes periodic banding whatever the orientation or period of the stripes, so every banding condition in our benchmark can be sent to it without determining which one is present. We therefore label a set of calibration images by the candidate that denoises each one best, rather than by the corruption that produced it. A new image then needs only to be matched to calibration images that the same denoiser served well, whether or not they share its corruption.}

\rev{To match images, we describe each one by thirteen statistics of its noise, estimated from the corrupted image alone. They measure properties of the process that produced the noise, which persist across scenes: its amplitude, the shape of its distribution, its dependence on the signal, and its spatial structure. A pool of frozen denoisers, classical or pretrained, is calibrated once by running every candidate on the calibration images and recording the best one for each, and the result can be shared and later extended with new references or new candidates. A new image is mapped into the descriptor space, matched to its nearest calibration image, and denoised by that image's best candidate, at a cost of one denoising call and no retraining.} \rev{In this space, an image's nearest neighbor has the same best denoiser in 83 to 89\% of cases but the same corruption in only 60\%, so matching by best denoiser succeeds far more often than matching by corruption would.}

On corruptions withheld from calibration, routing recovers 43 to 66\% of the improvement available to a perfect per-image choice, using a single denoising call, in the three pools where that improvement exceeds 2~dB, and it keeps that gain at severities above the calibrated range. In the fourth pool, where the available improvement is 0.93~dB, routing matches the best overall denoiser. \rev{With all 36 candidates in one pool, it gains 1.20~dB over the best overall denoiser. Running every candidate and keeping the output preferred by a no-reference quality score or by Monte-Carlo SURE} \citep{mittal2012making,ke2021musiq,ramani2008monte} \rev{costs one denoising call per candidate, and in every pool each of seven such selectors recovers less of the available improvement than routing. A classifier that first identifies the noise type and applies the matching denoiser} \citep{liu2020classification} \rev{also does worse, mainly on images containing two kinds of noise, which no single type describes. Learned routers trained on the same calibration labels, on our statistics or directly on the pixels, do no better than nearest-neighbor matching (Appendix~\ref{app:learned}). The gain therefore comes from labeling calibration images by their best denoiser, and once they are labeled this way a designed descriptor that requires no training is sufficient.}

On real camera noise, which no calibration condition reproduces, routing falls below the best overall denoiser, \rev{but the descriptor flags most such images as unlike any calibration image before a denoiser is run, and} adding labeled references of that noise restores the gain without retraining. \rev{JPEG compression is handled the same way: with references of compressed images and one new candidate designed for compression, routing comes within 0.25~dB of a perfect per-image choice (Appendix~\ref{app:new-noise}). Poisson--Gaussian noise, which no calibration condition contains, falls next to the Poisson conditions it combines and is routed to the denoisers that serve them. Because each statistic measures a known property of the noise, removing groups of statistics shows which properties decide the choice for which corruptions (Appendix~\ref{app:ablation}), and a failure can be traced to a property the descriptor does not measure and corrected by adding a statistic, without retraining any denoiser.}

\section{Related work}
\label{sec:related}

Prior methods for choosing a denoiser under an unknown corruption make the choice either after running the candidates or before, from a description of the corruption. Methods of the first kind score each output without a clean reference, using measures designed for denoising \citep{zhu2010automatic,lu2019noreference,rakhshanfar2018sparsity}, general no-reference quality scores \citep{mittal2012making,ke2021musiq}, or Monte-Carlo SURE \citep{ramani2008monte}, and they must produce and score every candidate output before deciding. Methods of the second kind classify the noise type and dispatch to a matched network \citep{liu2020classification,lemarchand2020noisebreaker}, estimate the parameters of a sensor noise model \citep{foi2008practical,huang2025beyond}, or route each input to a degradation-specific expert through a learned representation \citep{potlapalli2023promptir,ashrafee2026restoring}. In each case the corruption is described by its identity or by model parameters, and the denoiser follows from that description. A third line trains one network to cover many corruptions \citep{zhang2021plug,zamir2022restormer,zhang2023practical}; such networks \rev{are among the candidates of our pretrained} pool. We label calibration references by the candidate that denoises them best rather than by their corruption, so a query need only be matched to references served by the same denoiser, and our descriptor is designed rather than learned. Appendix~\ref{app:related} discusses each line of work in more detail.

\section{Method}
\label{sec:method}

Routing chooses a candidate denoiser for a new image by comparing its descriptor with the descriptors of a set of calibration images whose best performing denoiser is known. Its design therefore depends on which corruptions the descriptor must distinguish and on which candidates are available to choose from. We first define the corruption families and the properties along which they vary (Section~\ref{sec:corruptions}), then the descriptor built to measure those properties (Section~\ref{sec:descriptor}) and the pools of candidate denoisers (Section~\ref{sec:pools}). Section~\ref{sec:routing} describes the routing procedure, Section~\ref{sec:comparisons} the methods we compare it with, and Section~\ref{sec:protocol} the evaluation protocol.

\subsection{Corruptions}
\label{sec:corruptions}

We consider ten families of corruption, each with one to three physical parameters (Table~\ref{tab:corruptions} in Appendix~\ref{app:corruptions}). Every family has a parameter that sets the magnitude of the corruption. Beyond amplitude, the families are chosen to vary the other three properties the descriptor is meant to capture: the marginal law of the noise (Gaussian, Laplacian, uniform, and impulse), its dependence on the signal (Poisson and speckle), and its spatial structure (correlated, anisotropic, banding, and annular). Appendix~\ref{app:corruptions} defines each family and gives its physical motivation.
A \emph{condition} fixes a family and its parameter values, for example Gaussian noise with $\sigma = 25$ or banding with amplitude 20 and a period of 10 pixels. A \emph{mixture} assigns each pixel at random to one of two conditions, so the image contains both noises but no pixel receives both (Appendix~\ref{app:corruptions}).

\subsection{Descriptor}
\label{sec:descriptor}

The descriptor, which we call \Dxiii, maps a corrupted image to thirteen statistics, each chosen to measure one of the properties along which the corruption families of Section~\ref{sec:corruptions} vary. The statistics are computed from the corrupted image alone and are designed to respond more strongly to the corruption than to the scene. Figure~\ref{fig:sensitivity} in Appendix~\ref{app:statistics} shows, for each statistic, the range of its values as one corruption parameter is varied, relative to its variation across photographs and noise realizations.

\paragraph{Statistics.} Table~\ref{tab:statistics} in Appendix~\ref{app:statistics} groups the thirteen statistics by the property each measures. All are computed on a high-pass residual of the image, formed by splitting it into the four interleaved phases of a $2\times2$ decomposition and subtracting adjacent phases, giving two horizontal and two vertical differences. The decomposition follows the pixel-shuffle downsampling of \citet{lee2022apbsn}. Differencing adjacent phases suppresses image content that is shared locally while retaining the statistics of the corruption. The spectral statistics, except for the energy in four frequency bands, describe how concentrated and how directional the spectrum is without recording the frequency at which the energy lies. A periodic corruption whose frequency changes therefore stays close in descriptor space, and such a change leaves the denoising problem largely the same.

\paragraph{Calibration set and standardization.} Because the variance and the kurtosis of the residual range over several orders of magnitude, we replace the variance by its logarithm and the kurtosis by its inverse hyperbolic sine (Appendix~\ref{app:statistics}). We then standardize each statistic to zero mean and unit variance over a \emph{calibration set}, consisting of clean photographs corrupted under a fixed collection of calibration conditions (Table~\ref{tab:calibration} in Appendix~\ref{app:benchmarks}).

\subsection{Pools of denoisers}
\label{sec:pools}

A \emph{candidate} is a denoiser that is applied without retraining or tuning, in one call per image, and a \emph{pool} is a fixed set of candidates. Routing can improve on a single denoiser only when the candidates in a pool are strong on different corruptions. We therefore evaluated four pools that differ in how strongly their candidates specialize.

The \emph{classical} pool contains 10 fixed operators, including Gaussian, median, bilateral, and Wiener filters, total-variation and a notch filter for periodic corruption. 
The \emph{specialist} pool contains 14 networks of a common architecture, a reduced U-Net \citep{ronneberger2015unet}, each trained on one calibration condition with photographs from DIV2K \citep{agustsson2017ntire}, none of which appear in our benchmarks. 
The \emph{\rev{pretrained}} pool contains \rev{8 networks released by other authors}, among them DRUNet \citep{zhang2021plug}, SwinIR \citep{liang2021swinir}, Restormer \citep{zamir2022restormer}, and SCUNet \citep{zhang2023practical}. Their training covered only Gaussian noise or real camera noise, and they are expected to perform more alike across corruptions.
\rev{The \emph{hybrid} pool contains 17 candidates: the eight networks of the pretrained pool, three other released networks, and six classical operators.}
Appendix~\ref{app:pools} lists the candidates in each pool and gives implementation details.

\begin{figure}[t]
    \centering
    \includegraphics[width=1\linewidth]{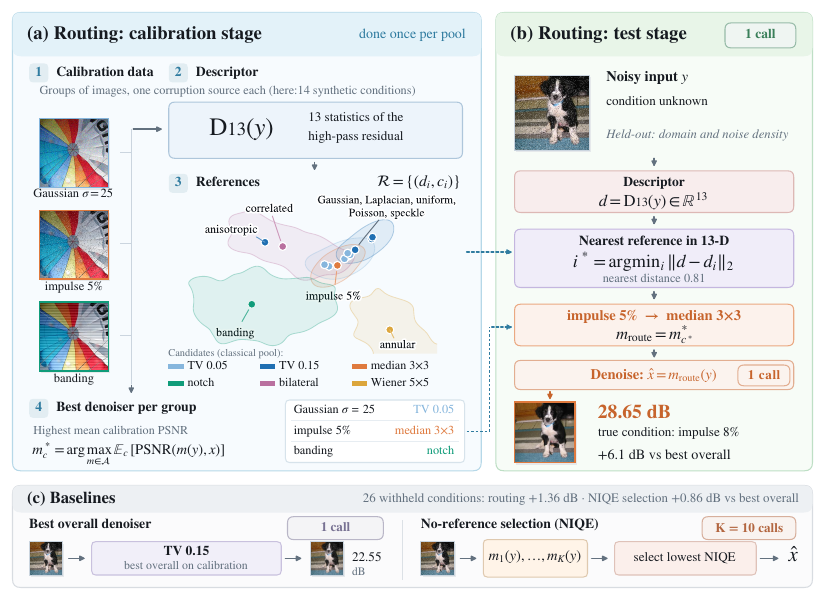}
    \caption{
    \textbf{Routing chooses a denoiser from the corrupted image before running any.} (a)~During calibration, candidates are run on groups of photographs that share a source corruption, and each calibration image becomes a reference in descriptor space (drawn schematically), paired with the candidate that denoised its group best. The descriptor, \Dxiii, consists of thirteen statistics that respond to the corruption more than to the scene.
    (b)~During test, a new image arrives without information about its corruption. It is matched to its nearest reference, and only that reference's candidate is run. The example is corrupted by impulse noise at a density and in a content domain not used in calibration; routing matches it to the reference for 5\% impulse noise and runs the median filter, which improves on the best overall denoiser by 6.1~dB. 
    (c)~We compare routing with two baselines: the best overall denoiser is chosen during calibration and run on every image; no-reference selection runs every candidate and keeps the output with the best quality score. With the classical pool, routing gains 1.36~dB over the best overall denoiser on withheld conditions, more than the 0.86~dB gained by selection with NIQE, which runs ten candidates instead of one. TV denotes total-variation denoising.
    }
    \label{fig:pipeline}
\end{figure}

\subsection{Routing}
\label{sec:routing}

Our routing procedure has a calibration stage, carried out once for each pool, and a test stage, carried out for each new image (Figure~\ref{fig:pipeline}). In the calibration stage, every candidate $m$ in the pool $\mathcal{A}$ is run on every image of the calibration set. Writing $y$ for a corrupted image, $x$ for its clean photograph, and $\mathbb{E}_c$ for the mean over the images of condition $c$, we label each calibration condition with the candidate of highest mean PSNR, $m_c^* = \arg\max_{m\in\mathcal{A}} \mathbb{E}_c[\mathrm{PSNR}(m(y),x)]$. Each calibration image $i$ then serves as a reference $(d_i, c_i)$, consisting of its standardized descriptor and its condition.

In the test stage, the standardized descriptor $d$ of a new image $y$ is computed from $D_{13}(y)$, the nearest reference $i^* = \arg\min_i \|d - d_i\|_2$ is retrieved, and the candidate $m_{\mathrm{route}} = m^*_{c_{i^*}}$ is run on the image. Routing therefore costs one descriptor computation and one denoising call. Computing the descriptor takes 23 ms for a $256\times256$ image on one core of an Apple M2~Max, comparable to a median or total-variation filter. Each calibration image serves as its own reference. We do not average the descriptors of a condition into one prototype per condition, because images of the same condition can fall in separate regions of the space.

Because labels record which candidate performs best and not which corruption is present, the procedure does not need to identify the corruption of a new image. As noted in the introduction, an image matched to the wrong condition is still denoised well when that condition has the same or a similarly effective best candidate.

\subsection{Comparisons}
\label{sec:comparisons}

We compare routing with three other ways of choosing a candidate denoiser for each test image. All comparisons are made within a pool, using only its candidates, and PSNR is always measured against the clean image. The three differ in how many candidates they run and in what information they use to choose among them.

The \emph{best overall} denoiser is the candidate with the highest mean PSNR on the calibration set. It is chosen once and run on every test image, so, like routing, it costs one call per image. It is the alternative named in the introduction, and we report all gains as the difference in PSNR from it.

The \emph{oracle} runs every candidate on a test image and keeps the output with the highest PSNR. Because it needs the clean image, it cannot be used in a realistic scenario, and we use it only to measure the largest gain that any choice among a pool's candidates can achieve. Its difference from the best overall denoiser is the \emph{oracle gap}, and we report the fraction of this gap that routing recovers.

\emph{No-reference selection} also runs every candidate, but it keeps the output that scores best under a no-reference image quality metric, computed on the output alone. Unlike the oracle, it can be used in the absence of ground truth, but it chooses correctly to the extent that the metric orders the outputs as PSNR would. It costs at least one call per candidate and one quality-score evaluation per output, whereas routing costs one call in total.

\subsection{Evaluation protocol}
\label{sec:protocol}

All photographs came from ten content domains, the ten classes of ImageNette \citep{howard2019imagenette}, with thirty photographs per domain in each benchmark (Appendix~\ref{app:benchmarks}). To keep calibration and test images separate, we held out one domain at a time, calibrated on photographs from the other nine, and tested on photographs from the held-out domain. Results were pooled over the ten held-out domains.
For each held-out domain, the references, the standardization statistics, the labels of the calibration conditions, and the best overall denoiser were all recomputed from the other nine domains.

We evaluated routing on two benchmarks, one testing transfer to conditions and mixtures not used in calibration, the other testing transfer to severities not used in calibration within a family.% (Table~\ref{tab:experiments}). 
The \emph{discrete} benchmark used 40 conditions, of which 14 formed the calibration set (Section~\ref{sec:descriptor}) and the other 26 were withheld. The withheld conditions included parameter values not used in calibration, variants of the structured families, and mixtures. Routing was tested only on the withheld conditions.

The \emph{parametric} benchmark varied one physical parameter per family, holding the others fixed, at six severity levels numbered 1 to 6 from lowest to highest, and used its own calibration set. Three splits divided the levels between calibration and test: S1 calibrated on levels 1, 3, and 5 and tested on levels 2 and 4, S2 calibrated on levels 1 to 3 and tested on levels 4 to 6, and S3 calibrated on levels 4 to 6 and tested on levels 1 to 3 (Appendix~\ref{app:benchmarks}).

We report the mean gain in PSNR over test images, where a test image is one photograph corrupted under one condition, and every test image has equal weight. Confidence intervals are 95\% intervals from a paired bootstrap over content domains and photographs, with 
$5{,}000$ replicates (Appendix~\ref{app:benchmarks}). Each photograph was corrupted once under each condition, with an independent noise realization, so every condition is evaluated over 300 realizations and none is repeated under a second seed.

\section{Results}
\label{sec:results}

The gain of routing over the best overall denoiser depends on the descriptor, which must place conditions that share a best candidate near each other, and on the pool, whose candidates must differ enough on those conditions for the choice to matter. We first test whether images that are close in descriptor space call for the same denoiser (Section~\ref{sec:res-descriptor}). We then measure the gain of routing on the conditions withheld from calibration (Section~\ref{sec:discrete}) and break it down by corruption family (Section~\ref{sec:families}). Finally, we test routing at severities not used in calibration (Section~\ref{sec:parametric}) and on real camera noise, which no calibration condition reproduces (Section~\ref{sec:sidd}).

\subsection{The descriptor space}
\label{sec:res-descriptor}
\begin{figure}[t]
\centering
\includegraphics[width=0.75\linewidth]{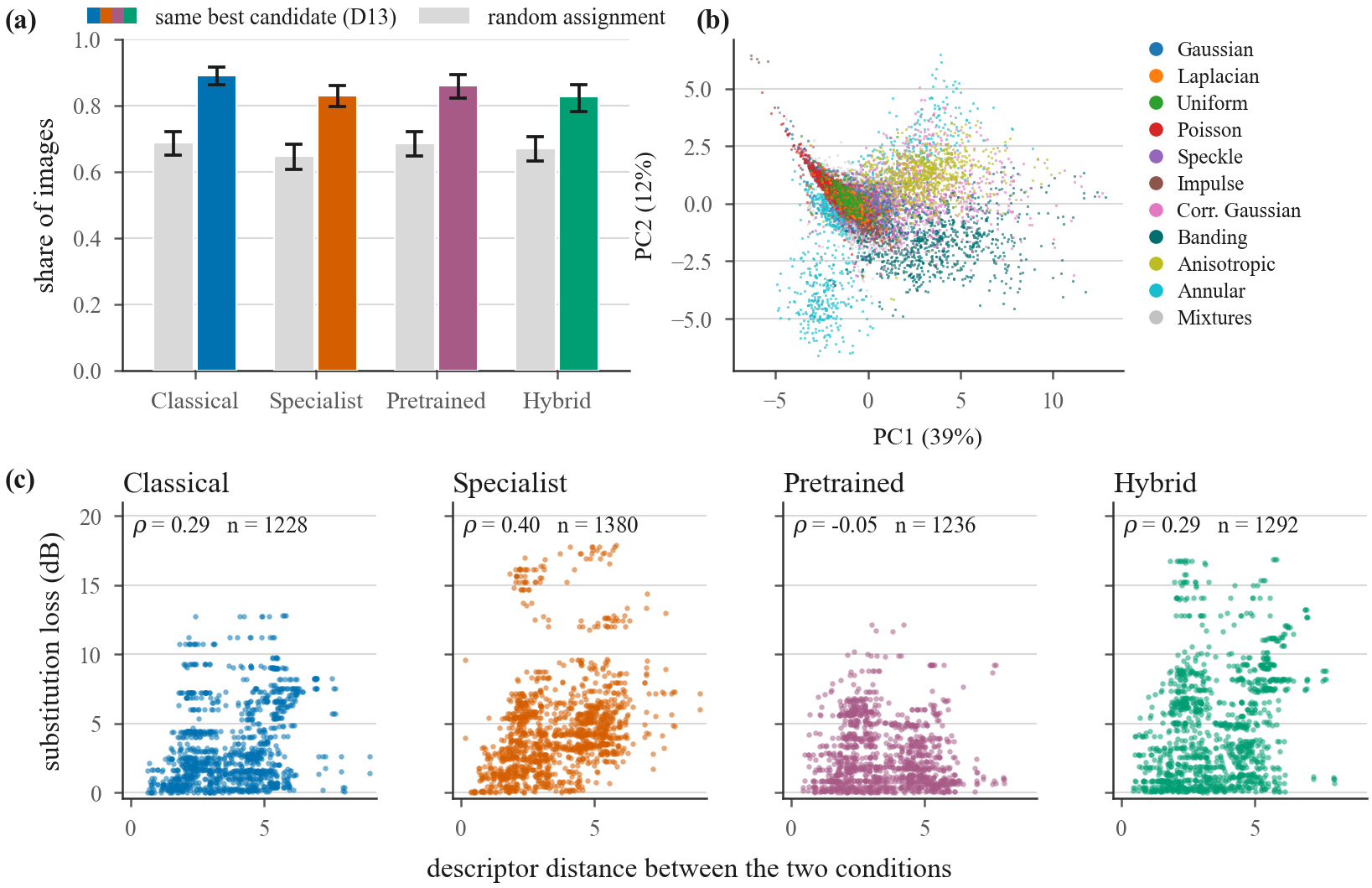}
\caption{\textbf{\Dxiii places noise conditions that call for the same denoiser near each other.} (a)~Fraction of images whose nearest neighbor under \Dxiii, retrieved over all 40 conditions, had the same best candidate as the query (colored bars). Gray bars show the expected fraction when best candidates are assigned to conditions at random, with each candidate remaining best on the same number of conditions. Error bars are 95\% paired-bootstrap intervals.
(b)~Descriptors of 12{,}000 images of the discrete benchmark projected onto their first two principal components, which account for 39\% and 12\% of the variance, colored by noise family. Mixtures of two conditions are shown in gray. Families that overlap in this projection may be separated along the remaining components.
(c)~Substitution loss against descriptor distance for each ordered pair of conditions. The substitution loss is the drop in PSNR when images of the first condition are denoised by the best candidate of the second. Pairs that share a best candidate have zero loss and are omitted from the plot and the correlations.}
\label{fig:descriptor-dists}
\end{figure}

Routing can only be as good as the descriptor space it relies on. Before measuring its gain, we therefore tested whether images that are close in this space call for the same denoiser.
We tested this on the discrete benchmark by retrieving, for each image, its nearest neighbor among the images of all 40 conditions from the other nine content domains. The neighbor had the same best candidate denoiser as the query in 83 to 89\% of images, depending on the pool, even though it had the same noise condition in only 60\% (Figure~\ref{fig:descriptor-dists}a). Because a few candidates are best on many conditions, neighbors agree to some extent by chance: assigning best candidates to conditions at random, keeping the number of conditions each candidate wins, gave 65 to 69\%. The descriptor therefore agrees far more often than the random assignment does. Routing can send an image to the right denoiser even when it matches the image to the wrong condition.

When the neighbor's best candidate differed from the query's, the loss depended on how well the neighbor's denoiser served the query. For each ordered pair of conditions, we denoised the images of the first condition with the best candidate of the second and measured the mean drop in PSNR relative to the first condition's own best candidate, which we call the substitution loss. We measured the distance between two conditions as the Euclidean distance between the means of their descriptors. Among pairs whose best candidates differed, this substitution loss was positively rank-correlated with descriptor distance in the classical, specialist, and hybrid pools (Spearman correlations of 0.29, 0.40, and 0.29; Figure~\ref{fig:descriptor-dists}c). Routing sends a query to its nearest reference, so its errors fall between conditions that are close in this space and tend to lose less PSNR than errors between distant conditions. In the pretrained pool there was no relation ($-0.05$), a case we return to in Section~\ref{sec:discrete}.

\subsection{Routing on withheld conditions}
\label{sec:discrete}

\begin{figure}[t]
    \centering
    \includegraphics[width=1\linewidth]{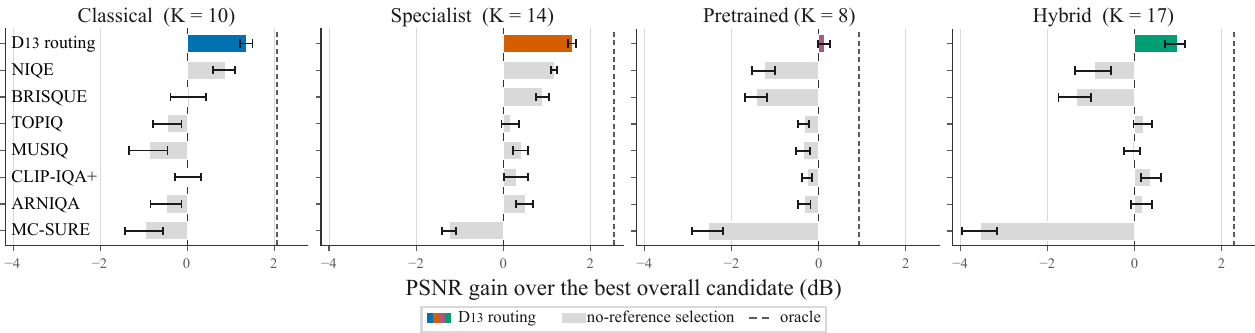}
    \caption{\textbf{Selection from the noisy image versus selection from denoised outputs.} Gain in PSNR over the best overall candidate for each pool. Colored bars show routing with \Dxiii, which runs one candidate. Gray bars show selection among the outputs of all $K$ candidates by a no-reference quality score, or by MC-SURE, which requires $2K$ calls. Dashed lines mark the oracle gap (gain available to a per-image choice with access to the clean target). Error bars are 95\% paired-bootstrap intervals. \Dxiii routing exceeds every no-reference selector in every pool.}
    \label{fig:forest}
\end{figure}

Choosing the right denoiser for an image is not enough to improve on the best overall denoiser, because on many images the two are the same candidate and routing then runs it. A gain can arise only on images where the two choices differ. We measured it on the 26 conditions of the discrete benchmark withheld from calibration, 14 of which are mixtures.
The gain of routing over the best overall denoiser was 1.36~dB for the classical pool, 1.58~dB for the specialist pool, and 0.99~dB for the hybrid pool, with 95\% intervals lying entirely above zero (Figure~\ref{fig:forest}). These gains recovered 66, 62, and 43\% of the oracle gap. For the pretrained pool the gain was 0.14~dB, recovering 15\% of a gap of 0.93~dB. The 95\% interval runs from $-0.01$ to 0.27~dB, so routing matched the best overall denoiser but did not measurably improve on it. Both the pool and the descriptor limit routing here: the oracle gap is the smallest of the four pools, and descriptor distance did not predict substitution loss (Section~\ref{sec:res-descriptor}).

\begin{figure}[t]
    \centering
    \includegraphics[width=.87\linewidth]{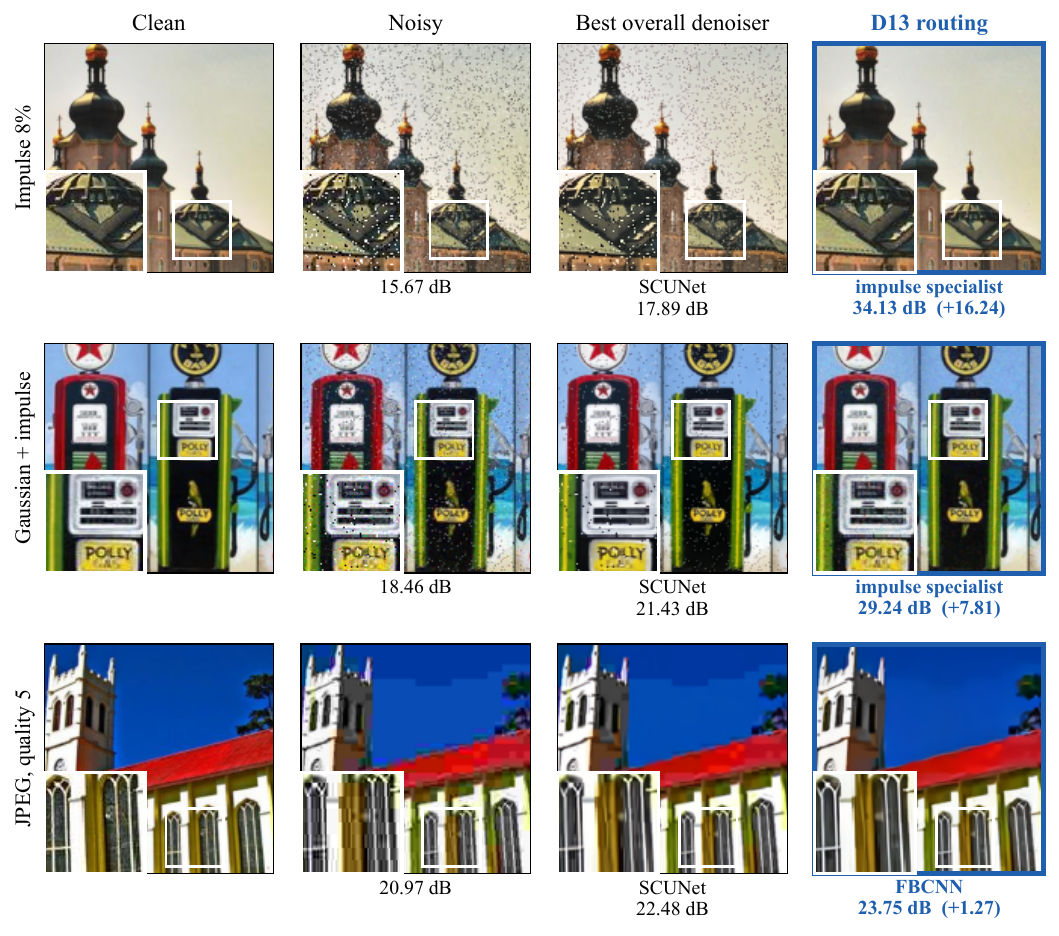}
    \caption{\rev{\textbf{Examples of routing over all 36 denoisers.} Each row shows one test image and
    corruption. Columns show the clean image, the corrupted image, the output of the best overall
    denoiser of the 36 (SCUNet, chosen on the calibration images in every fold), and the output of
    the candidate that routing selected. PSNR is given under each panel, with the gain of routing
    over SCUNet in parentheses, and the insets enlarge the region in the white box.
    \emph{Top:} impulse noise at 8\%, a density not used in calibration. Routing selects the
    specialist network trained on 5\% impulse noise, while SCUNet leaves most of the impulses in place.
    \emph{Middle:} a mixture withheld from calibration, in which 30\% of the pixels receive Gaussian noise ($\sigma = 25$) and 70\% receive impulse noise at 5\%. Routing again selects the impulse specialist.
    \emph{Bottom:} JPEG compression at quality 5, a corruption absent from the benchmark. This row
    uses the setting of Appendix~\ref{app:new-noise}: labeled JPEG examples from the other nine content domains are added to the calibration set, and FBCNN, a network designed for compression artifacts, is added
    as a 37th candidate. SCUNet remains the best overall denoiser. These examples were chosen to illustrate large gains; Section~\ref{sec:families} reports the gains by family, including those on which routing gained little or lost PSNR.}}
    \label{fig:example}
\end{figure}

\paragraph{All candidates in one pool.} We also routed over the 36 distinct candidates of the four pools together. The best overall denoiser of this pool is \texttt{S\_SCUNet\_real}, and routing gained 1.20~dB over it (95\% interval $[0.98, 1.36]$), recovering 39\% of an oracle gap of 3.06~dB. The routed PSNR was 27.68~dB, higher than in any of the four pools, so the errors introduced by a larger pool were smaller than the gain from the additional candidates. Removing the classical operators left the routed PSNR unchanged, and removing either the specialist or the pretrained networks lowered it by about 0.2~dB (Appendix~\ref{app:merged}). Figure~\ref{fig:example} shows three images on which routing over the 36 candidates chose a better denoiser than \texttt{S\_SCUNet\_real}.

\paragraph{Comparison with no-reference selection.} Routing chooses a denoiser before running any. The alternative is to run every candidate and choose among the outputs, using a score computed without the clean image. We compared routing with seven such selectors (Appendix~\ref{app:selectors}). Six are no-reference quality scores: NIQE \citep{mittal2012making}, BRISQUE \citep{mittal2012brisque}, and four recent learned scores, MUSIQ \citep{ke2021musiq}, CLIP-IQA+ \citep{wang2023clipiqa}, TOPIQ \citep{chen2024topiq}, and ARNIQA \citep{agnolucci2024arniqa}. The seventh, Monte-Carlo SURE \citep{ramani2008monte}, estimates the mean squared error of each output under the assumption of additive white Gaussian noise, with the noise level estimated from the corrupted image, and requires a second denoising call per candidate.
Routing exceeded every selector in every pool, with 95\% intervals of the difference lying entirely above zero, at one denoising call against $K$ for the quality scores and $2K$ for SURE. The closest selector, NIQE, fell short of routing by 0.41~dB in the specialist pool and 0.50~dB in the classical pool. In the pretrained pool, where the candidates differ least, every selector did worse than running the best overall denoiser on every image, whereas routing did not. SURE estimates the error that PSNR measures, but only under its Gaussian assumption; on impulse and correlated noise its ranking of the candidates was negatively correlated with their true error (Appendix~\ref{app:selectors}).

\paragraph{Comparison with noise classification.} The other alternative is to identify the corruption first. We reimplemented the classifier of \citet{liu2020classification}, which assigns an image to one of four noise types, and compared it with routing restricted to the same types, on the eight withheld conditions composed only of them (Appendix~\ref{app:liu}). With the published preprocessing, routing gave higher PSNR than the classifier in every pool, by 0.47 to 0.70~dB. At full resolution, the classifier still gave lower PSNR in three pools and did not differ in the fourth. The difference arose on mixtures of two types, for which the classifier has no label.

\subsection{Routing at unseen severities}
\label{sec:parametric}

The withheld conditions of the discrete benchmark differ from the calibration conditions in several parameters at once. The parametric benchmark varies one parameter per family and tests severities between calibrated ones (S1), above the calibrated range (S2), and below it (S3) (Section~\ref{sec:protocol}). Routing kept its gain in S1 and S2, recovering about three quarters of the oracle gap in the classical and hybrid pools and about half in the pretrained pool. In S3 it recovered only a quarter to a third, although the oracle gap was largest there (Figure~\ref{fig:parametric} in Appendix~\ref{app:parametric}). Mild noise changes the statistics the descriptor measures little and is hard to distinguish from image content, so severities below the calibrated range are matched to the mildest calibrated reference, whose best candidate over-smooths them. The specialist pool is not evaluated on this benchmark, because its networks were each trained on one calibration condition of the discrete benchmark (Appendix~\ref{app:parametric}).

\subsection{Gain by corruption family}
\label{sec:families}

The gains of Section~\ref{sec:discrete} are averages over corruption families whose oracle gaps differ by more than two orders of magnitude, from 0.04 to 15.8~dB (Table~\ref{tab:families} in Appendix~\ref{app:families}). Routing recovered nearly all of the gap where it was largest. On impulse noise it recovered 99\% in every pool, and on banding 81 and 80\% in the classical and hybrid pools, where the gap was about 8~dB. These two families form 3 of the 26 withheld conditions but accounted for 0.76 of the 1.36~dB gained by the classical pool, and for all of the gain of the hybrid and pretrained pools. On the other 23 conditions, routing gained 0.69~dB with the classical pool (95\% interval $[0.51, 0.84]$) and 1.09~dB with the specialist pool ($[0.99, 1.17]$), whereas with the hybrid and pretrained pools the gain was $-0.06$~dB, with intervals that included zero.
Where the oracle gap was small, routing could lose PSNR. In the pretrained and hybrid pools, the gap on correlated, annular, and signal-dependent noise was below 0.2~dB, yet routing lost 0.47 to 1.13~dB on these families by selecting candidates worse than the best overall denoiser. The classical pool did the same on additive noise, losing 0.76~dB where the gap was 0.20~dB. The largest shortfall on a family with a large gap was banding in the specialist pool, where routing recovered 7\% of a 3.0~dB gap. This pool contains a single banding specialist, trained on horizontal stripes, and the withheld banding condition has vertical ones.
Mixtures, which form 14 of the 26 withheld conditions, combine two families in one image. The classical and specialist pools recovered 59 and 58\% of their gap on mixtures, which was about 1.6~dB, and the pretrained and hybrid pools recovered 13 and 8\%.

\subsection{Noise outside calibration}
\label{sec:sidd}

Routing runs the candidate that was best at the nearest reference, so it can be no better than the references near a query. When a query's noise resembles no calibration condition, its nearest reference does not describe it, and routing should lose its advantage over the best overall denoiser and may fall below it. The remedy the method suggests is to add references of the new noise, labeled with their best candidate, which requires clean images of that noise but no retraining. We tested both expectations on real camera noise, which combines dependence on the signal, spatial correlation, and the effects of the camera's processing, and which no calibration condition reproduces.

\begin{figure}[h]
\centering
\includegraphics[width=0.6\linewidth]{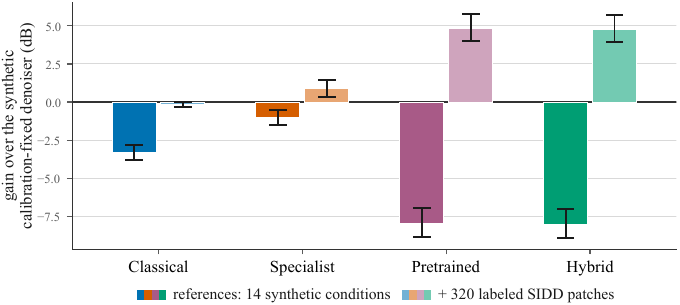}
\caption{\textbf{Adding references repairs routing on real camera noise.} Gain in PSNR on the SIDD validation images over the best overall denoiser of the synthetic calibration set, with references from that set alone and with 320 labeled SIDD Small patches added to it. Error bars are 95\% intervals.}
\label{fig:sidd}
\end{figure}

We routed the 1{,}280 sRGB validation blocks of SIDD \citep{abdelhamed2018high} with all four pools. With the synthetic calibration set alone, the median distance from a SIDD block to its nearest reference was 3.42, against 1.02 for images of withheld synthetic conditions and 0.81 for images of calibration conditions. As expected, routing did worse than the best overall denoiser of the synthetic calibration set, by 1.01~dB with the specialist pool, 3.29~dB with the classical pool, and about 8~dB with the pretrained and hybrid pools, where most of the loss was the cost of the fixed choice itself rather than of routing (Figure~\ref{fig:sidd}, Appendix~\ref{app:sidd}).

The descriptor also indicated, before any denoiser was run, when a query lay outside the calibration conditions. We flagged the images whose descriptor had a negative log-likelihood, under the best-fitting calibration condition, above the 99th percentile of the calibration images. The flag fired on 90\% of the SIDD blocks and on 7\% of the images of withheld synthetic conditions. On those synthetic images, routing recovered a smaller share of the oracle gap when the image was flagged, in all four pools (Appendix~\ref{app:sidd}).

We then added the 320 SIDD Small patches as references, each labeled with the best candidate for its ISO band. Routing then recovered most of the loss in every pool, and changed the results on withheld synthetic conditions by at most 0.04~dB. Part of the recovery was a change of default rather than a per-image choice, since the real references point to a network trained on real camera noise; in the pretrained and hybrid pools that network accounted for the whole of the gain, and a per-image choice could have added at most 0.12 and 0.27~dB, respectively. In the specialist pool, whose networks had never seen real noise, routing gained 1.84~dB [1.11, 2.59] over the best candidate on the real references, out of a gap of 3.48~dB (Appendix~\ref{app:sidd}).

\section{Discussion}

\rev{Labeling calibration images by their best denoiser lets a pool of existing denoisers adapt to the noise of each image, without retraining any candidate and at the cost of one denoising call. Routing helped most when the candidates were strong on different corruptions, and matched the best overall denoiser when they were all trained for the same noise.} Within a family, routing transferred to severities above the calibrated range nearly as well as between calibrated severities, but less so to milder noise, so the calibration conditions should include the mildest noise a user expects.

\paragraph{Limitations.} Routing depends on references that resemble the noise of a new image, and covering a new kind of noise requires clean images of it for labeling. Where little gain was available from a per-image choice, routing sometimes selected a worse candidate than the best overall denoiser. The available gain can be measured on the calibration set before routing is used, and images flagged as uncovered (Section~\ref{sec:sidd}) can be sent to the best overall denoiser. The descriptor does not record properties it was not designed to measure, such as the orientation of stripes. With all 36 candidates, the specialist trained on horizontal stripes outperformed the notch filter on the horizontal banding of the calibration set and became its label, so vertical banding was matched to that reference and sent to the specialist rather than to the notch filter. Both the labels and the evaluation used PSNR, and a perceptual metric may prefer a different candidate; labeling the references by another metric requires no change to the procedure. Two natural extensions are routing regions of an image separately and choosing the denoiser at each iteration of a plug-and-play method.

\paragraph{Extension to new noise.} The procedure can be extended to a kind of noise it was not calibrated for by adding labeled images of that noise as references and, where no candidate suits it, a denoiser designed for it. For real camera noise, which lay far from the synthetic references, 320 such images removed most of the loss without retraining any candidate, and changed the results on synthetic noise by at most 0.04~dB. For JPEG compression, references together with FBCNN brought routing within 0.25~dB of a perfect per-image choice; the gain on the withheld synthetic conditions fell by 0.26~dB, because 7.8\% of their images were matched to a JPEG reference (Appendix~\ref{app:new-noise}). Whether a pool is worth routing over can be assessed in advance from the gain available on the calibration set, and whether a new image is covered by the references from the flag of Section~\ref{sec:sidd}.

\subsection*{Acknowledgments}
L.D. was supported by the European Union's Marie Sk\l{}odowska-Curie Action under grant agreement no.~101207931. Generative AI tools were used to implement parts of the code, to revise figures, to edit the text for clarity, and to search the literature. The authors reviewed and verified all AI-assisted work and take full responsibility for the content of this article.

\subsection*{Reproducibility statement}

All data are public. The photographs come from ImageNette and DIV2K, and the real camera noise from the SIDD validation set (Appendices~\ref{app:benchmarks} and~\ref{app:sidd}). Appendix~\ref{app:corruptions} defines the corruption families and their parameters, and Appendix~\ref{app:statistics} defines the thirteen statistics of the descriptor. Appendix~\ref{app:pools} lists every candidate with its implementation, checkpoint, and software version, and describes the training of the specialist networks. Appendix~\ref{app:benchmarks} gives the benchmarks and the bootstrap procedure, and Appendices~\ref{app:selectors} and~\ref{app:liu} give the baselines. Code is available at \url{https://github.com/Giovanni-Nocerino/blind-denoising}.

\bibliographystyle{iclr2027_conference}
\bibliography{refs}

\appendix

\section{Extended related work}
\label{app:related}

This appendix expands Section~\ref{sec:related}. Prior work either defers the choice of denoiser until the candidates have been run and their outputs can be compared, or makes it in advance by describing the corruption, as a type, as the parameters of a noise model, or as a learned embedding. A third line avoids the choice by training one network to cover every corruption. None of these methods asks what a description of the noise must preserve in order to choose among given denoisers. Our descriptor is designed around that question.

\paragraph{Selecting a denoiser from its output.} One line of work runs the candidates and judges the results. \citet{zhu2010automatic} score a denoised image by the coherence of its gradients, \citet{lu2019noreference} learns a regression from features of the denoised image and of the removed noise, and \citet{rakhshanfar2018sparsity} score a result by the structure left in the residual. A second group scores the outputs with a general no-reference quality measure rather than one designed for denoising, from the natural-scene statistics of NIQE \citep{mittal2012making} and BRISQUE \citep{mittal2012brisque} to learned scores such as MUSIQ \citep{ke2021musiq}, CLIP-IQA+ \citep{wang2023clipiqa}, TOPIQ \citep{chen2024topiq}, and ARNIQA \citep{agnolucci2024arniqa}. Monte-Carlo SURE \citep{ramani2008monte} instead estimates the mean squared error of each output, at the cost of a second denoising call per candidate and under an assumption of additive white Gaussian noise. All of these methods need no clean image and can tune a parameter as well as choose an algorithm, but they must produce every candidate output before deciding, and each output must then be scored; \citet{lu2019noreference} reports about five seconds per image for this step.

\paragraph{Recognizing the corruption.} A second line identifies the noise and applies a denoiser matched to it. \citet{liu2020classification} classify the noise type with one network and denoise with another, and \citet{lemarchand2020noisebreaker} classify the noise into five types, each with its own trained network, restoring the image in several passes. Classification is also used within a single method, to select the data term of a variational reconstruction \citep{fantasia2025patch} or to pretrain a restoration network \citep{hu2025universal}. These methods assume that the corruption must be identified for the denoiser to be chosen, and they train a classifier to identify it. Our descriptor measures properties of the corruption without training and uses them only to choose the denoiser.

\paragraph{One network for many corruptions.} A different response to the same problem is to train one network to cover a range of corruptions, whether by conditioning on an estimated noise level \citep{zhang2021plug}, by training blind over a range of severities \citep{zamir2022restormer}, or by synthesizing a wide degradation model at training time \citep{zhang2023practical}. \rev{Several of these networks are candidates of our pretrained pool.} Training of this kind makes the candidates of a pool perform alike, which is the case in which a choice among them has least to offer, and it requires the corruptions to be anticipated during training, whereas a pool can be extended by adding a denoiser and its calibration references.

\paragraph{Estimating noise parameters.} Rather than a discrete type, the noise can be described by the parameters of a sensor model. \citet{foi2008practical} fit the variance of smooth patches against their intensity, and \citet{huang2025beyond} estimate the two parameters of a Poisson--Gaussian model from a single raw image and condition a trained transformer on them. Our signal-dependence statistic is an estimate of this kind, computed on pixel differences rather than on patches selected for smoothness. Two scalars of a sensor model cannot express impulse, periodic, or spatially correlated corruption, which is where the largest part of our gain arises.

\paragraph{Routing to trained experts.} All-in-one restoration methods adapt a shared model to unknown degradations using learned degradation representations or prompts \citep{li2022all,potlapalli2023promptir}. \rev{Mixture-of-experts restoration models such as MEASNet \citep{yu2025measnet} and MoCE-IR \citep{zamfir2025complexity} select among experts inside one network, trained jointly with the router on a fixed set of degradations, whereas routing selects among independently developed denoisers that remain frozen (Appendix~\ref{app:learned}).} Closest to our setting, \citet{ashrafee2026restoring} freeze a pretrained backbone, train one adapter per degradation, and route each input to the nearest degradation prototype, separating degradations as distinct as noise, rain, and haze and scoring the routing against the degradation label. These approaches organize routing around degradation identity. We instead label references by the candidate that denoises them best, and we score routing by the PSNR of its output, using the corruption label only to analyze the descriptor space. The corruptions we distinguish are all noise, and two different corruptions can call for the same denoiser.

\section{Corruption families}
\label{app:corruptions}

Standard treatments of additive, impulse, and periodic noise are given by \citet{gonzalez2018digital}; signal-dependent noise in camera sensors is modeled by \citet{foi2008practical}, and speckle by \citet{goodman2007speckle}.
Intensities are on $[0, 255]$ unless stated otherwise. We write $x$ for the clean image, $y$ for the corrupted image, and $n$ for a noise field.

\begin{table}[h]
\centering
\caption{Corruption families and their parameters.}
\label{tab:corruptions}
\begin{tabular}{lll}
\toprule
Property varied & Family & Parameters \\
\midrule
Marginal law & Gaussian, Laplacian, uniform & $\sigma$ \\
             & Impulse & fraction of pixels $p$ \\
Signal dependence & Poisson & peak count \\
                  & Speckle & $\sigma$ \\
Spatial structure & Correlated & $\sigma$, correlation length \\
                  & Anisotropic & $\sigma$, correlation length in $x$ and in $y$ \\
                  & Banding & amplitude, period, orientation \\
                  & Annular & $\sigma$, center frequency $f_0$, bandwidth \\
\bottomrule
\end{tabular}
\end{table}

\paragraph{Marginal law.} Gaussian, Laplacian, and uniform noise add independent samples of standard deviation $\sigma$ to every pixel, $y = x + n$. Gaussian noise is the standard model of thermal and readout noise. Laplacian noise has heavier tails, and uniform noise, which approximates quantization error, has lighter ones. Impulse noise replaces a fraction $p$ of pixels, chosen independently, with extreme values, across the three RGB channels, as dead sensor elements or transmission errors do.

\paragraph{Signal dependence.} Poisson noise models photon counting. Intensity is mapped to a photon count by the fixed scale $\lambda = x\,\text{peak}/255$, so intensity 255 corresponds to the peak count in every photograph, each pixel is drawn from a Poisson distribution with mean $\lambda$, and the count is mapped back by the same scale; the noise variance is therefore proportional to intensity, and a lower peak gives noisier images. Speckle models coherent imaging, as in ultrasound and radar, and multiplies the image by a random factor, $y = x(1 + n)$ with $n$ of standard deviation $\sigma$, so the noise standard deviation grows in proportion to intensity.

\paragraph{Spatial structure.} Correlated noise is white Gaussian noise smoothed by a Gaussian kernel of standard deviation $\ell/2$ pixels, where $\ell$ is the correlation length, and then rescaled to standard deviation $\sigma$. It models the effect of interpolation and filtering in a camera's processing. Anisotropic noise uses the same filter with standard deviations $\ell_x/2$ and $\ell_y/2$ along the horizontal and vertical axes. Banding adds a sinusoidal pattern of horizontal or vertical stripes with a given amplitude and period, as produced by row- or column-dependent readout offsets and electrical interference. The phase of the pattern is drawn uniformly at random for each realization.
Annular noise multiplies the spectrum of white noise by a Gaussian ring, $\exp\bigl(-(\lVert f\rVert-f_0)^2/(2\,\mathrm{bw}^2)\bigr)$, and rescales the result to standard deviation $\sigma$. Here $f$ is the two-dimensional spatial frequency in cycles per pixel, with each component in $[-1/2,1/2)$, and $f_0$ is the center frequency of the ring; the calibration value $f_0=0.25$ is half the Nyquist frequency along each axis, a period of four pixels. The bandwidth $\mathrm{bw}$ is the standard deviation of the ring in amplitude, so the ring in power is narrower by $\sqrt{2}$. The ring depends on $\lVert f\rVert$ alone, which gives structure at a characteristic scale with no preferred direction.

\paragraph{Mixtures.} A mixture independently applies two conditions to the same clean image and subsequently selects between the two results on a pixel-by-pixel basis, using a random binary mask generated independently for each pixel and shared across the three color channels. Consequently, each pixel is subject to only one of the two types of noise and never both. The fourteen mixtures consist of seven pairs of conditions, each applied using two mixing fractions: $0.3$ (where 30\% of the pixels derive from the first condition of the pair and 70\% from the second) and $0.6$ (where the split is 60\% and 40\%).

\section{Statistics}
\label{app:statistics}

\begin{table}[h]
\centering
\caption{The thirteen statistics of \Dxiii, grouped by the property each one measures.
The first nine are computed on the four-phase residual of the color channels, the last four
on an estimate of the power spectrum of the luminance. The names are those used in
Appendix~\ref{app:d13-recipe}.}
\label{tab:statistics}
\begin{tabular}{@{}lll@{}}
\toprule
Property & Statistic & Name \\
\midrule
Amplitude           & variance                                              & \texttt{marg\_var} \\
Marginal law        & skewness                                              & \texttt{marg\_skew} \\
                    & excess kurtosis                                       & \texttt{marg\_kurt} \\
Signal dependence   & slope of the log variance across intensity bins       & \texttt{sig\_slope} \\
Spatial correlation & correlation at a lag of one pixel along rows                & \texttt{spat\_rlag01} \\
Radial spectrum     & energy in the four innermost of eight radial bands    & \texttt{freq\_band0-3} \\
\addlinespace
Spectral structure  & prominence of the largest peak                        & \texttt{peak\_prom} \\
                    & energy in the strongest half percent of bins          & \texttt{line\_frac} \\
                    & angular concentration at the peak radius              & \texttt{ang\_conc} \\
                    & anisotropy                                            & \texttt{anisotropy} \\
\bottomrule
\end{tabular}
\end{table}

\paragraph{Properties measured.} The four properties of Table~\ref{tab:statistics} are those along which the corruption families of Appendix~\ref{app:corruptions} vary, and each statistic is designed to respond to one of them. The variance of the residual measures amplitude. The skewness and the excess kurtosis measure the shape of the marginal law, separating Laplacian and impulse noise, which have heavy tails, from uniform noise, which has light ones. The slope of the log variance across intensity bins is zero for additive noise and grows with the rate at which the noise variance follows the signal, as it does for Poisson and speckle noise. The nine remaining statistics describe spatial structure: the lag-one correlation and the four radial bands respond to correlated and anisotropic noise, whose energy is concentrated at low frequencies, and the four structural statistics respond to banding and annular noise, which place their energy on a few frequencies or on a ring. Of the four, \texttt{peak\_prom} and \texttt{line\_frac} record how concentrated the spectrum is and \texttt{ang\_conc} and \texttt{anisotropy} how directional it is, none of them recording the frequency at which the energy lies.

\paragraph{Scaling.} Two of the thirteen statistics are transformed before standardization. The variance $v$ of the residual is replaced by $\log v$, and the excess kurtosis $\kappa$ by $\operatorname{asinh}\kappa = \ln\bigl(\kappa + \sqrt{\kappa^2+1}\bigr)$. Both range over several orders of magnitude across the corruption families, and in raw form a few conditions would dominate the Euclidean distance between descriptors. A logarithm cannot be used on the kurtosis, which is negative for light-tailed noise such as uniform noise; the inverse hyperbolic sine is nearly linear near zero and logarithmic for large values of either sign. Neither transformation has a fitted parameter. The remaining eleven statistics enter the descriptor as defined.

\subsection{\Dxiii implementation}
\label{app:d13-recipe}
This section specifies each statistic in enough detail for \Dxiii to be reimplemented. Throughout, $y$ denotes the corrupted image, with channels $y_c$, $c=1,\dots,C$ ($C=3$ for sRGB), on the intensity scale $[0,255]$. The first nine statistics are computed from a residual formed by differencing the four polyphase components of each channel, and the last four from an estimate of the noise power spectrum of the luminance. Each division, logarithm, and radius normalization is protected by a small positive constant. These constants do not change the statistics of a corrupted image, and we omit them below except where a reported value depends on one.

\paragraph{The residual.} Each channel is cropped to even side lengths, $H'=2\lfloor H/2\rfloor$ and $W'=2\lfloor W/2\rfloor$, and split into its four polyphase components,
\[
p^{c}_{ab}(i,j)=y_c(2i+a,\,2j+b),\qquad a,b\in\{0,1\},\quad 0\le i<H'/2,\ 0\le j<W'/2 .
\]
This is the pixel-shuffle downsampling that \citet{zhou2020awgn} and \citet{lee2022apbsn} use to break the spatial correlation of real sensor noise before it enters a network; here it is used to form differences. Of the six pairs of components, we difference the four that are adjacent, two along rows and two along columns,
\[
\mathcal{P}=\bigl\{(p_{00},p_{01}),\,(p_{10},p_{11}),\,(p_{00},p_{10}),\,(p_{01},p_{11})\bigr\},
\]
and for each channel and each pair $(p,p')\in\mathcal{P}$ we form a residual plane and a local-mean plane,
\[
r_k=\frac{p-p'}{\sqrt{2}},\qquad m_k=\frac{p+p'}{2},
\]
where $k$ indexes the $4C$ combinations of channel and pair.

The factor $1/\sqrt{2}$ gives the residual the variance of the noise. If $y=x+\eta$ with $\eta$ white of variance $\sigma^2$, the noise component of $r_k$ has variance $\sigma^2$, and so does $r_k$ wherever $x$ is locally constant. No other rescaling is applied, and intensities remain on $[0,255]$. A difference along rows has zero response at zero horizontal frequency, and a difference along columns at zero vertical frequency, so with both orientations the residual responds to noise at every frequency except the origin. Using both orientations also treats horizontal and vertical image edges alike. Cropping is the only treatment of the borders: each component has exactly $H'/2\times W'/2$ samples, and nothing is padded. Each pixel enters one difference along rows and one along columns, so for white noise the samples of a residual plane are independent. Statistics 1 to 9 treat the channels identically and do not combine them. The $4C$ planes are concatenated for the marginal and signal-dependence statistics (1--4), and the statistic is averaged over planes for the correlation and the radial bands (5--9). Only the four structural statistics (10--13) combine the channels, through the unweighted luminance $\ell=C^{-1}\sum_c y_c$.

\paragraph{Amplitude and marginal law (1--3).} Let $r$ denote the $4C$ residual planes concatenated into a single vector of length $N$, with mean $\bar r$ and variance $s^2=N^{-1}\sum_n(r_n-\bar r)^2$. The first three statistics are its variance, skewness, and excess kurtosis,
\[
\texttt{marg\_var}=s^2,\qquad
\texttt{marg\_skew}=\frac{1}{N}\sum_n\Bigl(\frac{r_n-\bar r}{s}\Bigr)^{3},\qquad
\texttt{marg\_kurt}=\frac{1}{N}\sum_n\Bigl(\frac{r_n-\bar r}{s}\Bigr)^{4}-3 .
\]
The excess kurtosis is zero when the noise is Gaussian, positive for heavier-tailed noise such as Laplacian and impulse noise, and negative for uniform noise.

\paragraph{Signal dependence (4).} The local-mean planes are concatenated in the same order as the residual, and their range $[\min m,\max m]$ is divided into $B=8$ bins of equal width. Let $\mathcal{B}_b$ be the set of residual samples whose local mean falls in bin $b$. The variance in each bin is
\[
v_b=\begin{cases}\widehat{\operatorname{Var}}\bigl(r_n : n\in\mathcal{B}_b\bigr) & \text{if } |\mathcal{B}_b|\ge 8,\\[2pt] s^2 & \text{otherwise,}\end{cases}
\]
and the statistic is the least-squares slope of $\log v_b$ against the bin index,
\[
\texttt{sig\_slope}=\frac{\sum_{b}(b-\bar b)\,(\log v_b-\overline{\log v})}{\sum_b (b-\bar b)^2},
\]
where $\bar b$ and $\overline{\log v}$ are means over the $B$ bins. Each residual sample is conditioned on the mean of the two pixels from which it was computed, so no window larger than the pair is involved. Because the bins span the intensity range of the image, the slope per unit intensity equals the statistic divided by the bin width, which differs between images.

The statistic is zero for additive noise and increases with the rate at which the noise variance grows with intensity. If the conditional variance follows a power law, $\operatorname{Var}(r\mid x)\propto x^{\gamma}$, then $\log v_b\approx\gamma\log x_b+\text{const}$, where $x_b$ is the center of bin $b$, and the fitted slope is $\gamma$ times a positive factor determined by the bin centers. For a given image the statistic is therefore proportional to $\gamma$, and a linear fit suffices without estimating the noise-level function. Poisson noise has $\gamma=1$ and speckle $\gamma=2$.

\paragraph{Spatial correlation (5).} For each residual plane, with $\tilde r_k=r_k-\bar r_k$, the correlation at a lag of one sample along rows is
\[
\rho_k=\frac{\bigl\langle \tilde r_k(i,j)\,\tilde r_k(i,j+1)\bigr\rangle}{\bigl\langle \tilde r_k(i,j)^2\bigr\rangle},
\qquad
\texttt{spat\_rlag01}=\frac{1}{4C}\sum_k \rho_k ,
\]
where $\langle\cdot\rangle$ averages over the positions at which both terms are defined. One sample of the residual lattice is two pixels of the image, and the same lag is taken on all four planes, so the coordinate is a fixed linear functional of the noise autocovariance along rows at that separation. It is zero for white noise, where every lag involved vanishes, and departs from zero when the noise carries horizontal dependence at that scale. 

\paragraph{Radial bands (6--9).} For each residual plane, the periodogram of the mean-removed plane is $P_k=\bigl|\mathcal{F}\{\tilde r_k\}\bigr|^2$, with zero frequency shifted to the center of the array. Write $\rho(u,v)$ for the distance of a bin from the center and $\rho_{\max}$ for the largest such distance on the grid, which is attained at a corner. The grid is divided into eight annuli of equal width in radius,
\[
\beta(u,v)=\min\Bigl(\bigl\lfloor 8\,\rho(u,v)/\rho_{\max}\bigr\rfloor,\,7\Bigr),
\]
the floor function being clipped so that the corner bins themselves fall in band 7. The fraction of the energy of plane $k$ in band $j$ is
\[
e^{(k)}_j=\frac{\sum_{\beta(u,v)=j}P_k(u,v)}{\sum_{u,v}P_k(u,v)},\qquad j=0,\dots,7,
\]
and the four statistics are the averages of the four innermost bands over planes, $\texttt{freq\_band}_j=(4C)^{-1}\sum_k e^{(k)}_j$ for $j=0,1,2,3$. The band edges follow from the grid alone and have no free parameter.

\paragraph{Why four bands.} The normalizing radius is the corner of the sampling square, $\sqrt{2}$ times the Nyquist radius, so band $j$ spans $[\,0.088\,j,\;0.088\,(j+1))$ cycles per sample of the residual lattice, against a Nyquist frequency of $0.5$. Bands 0 to 4 lie entirely inside the Nyquist disc, band 5 is cut by it, and bands 6 and 7 consist only of the corners of the square, so their content reflects the sampling geometry rather than the noise. That leaves five complete annuli, of which we keep the first four. Nothing is lost about how much energy sits above the fourth edge: since the eight fractions sum to one, the pooled share of bands 4 to 7 is one minus the sum of the four we keep, an affine function of them. The statistics of the following paragraph already describe narrow-band content at any frequency up to Nyquist, so the radial bands need only carry the coarse low-frequency profile, which is where the spatially structured families place their energy. Keeping four also matches the four structural statistics: with all eight bands, the radial block would carry close to half of the squared distance between standardized descriptors, the imbalance that the logarithm and the inverse hyperbolic sine of Section~\ref{sec:descriptor} are there to prevent.

\paragraph{The noise spectrum.} The last four statistics are computed from an estimate of the noise power spectrum of the luminance (not to be understood as a photometric luminance, as it is unweighted) $\ell=C^{-1}\sum_c y_c$, at full resolution rather than on the residual lattice. Three first differences are formed,
\[
d_1=\frac{\ell(i,j{+}1)-\ell(i,j)}{\sqrt{2}},\qquad
d_2=\frac{\ell(i{+}1,j)-\ell(i,j)}{\sqrt{2}},\qquad
d_3=\frac{\ell(i{+}1,j{+}1)-\ell(i,j)}{\sqrt{2}},
\]
each of which passes the noise spectrum $S$ through a known filter, with squared transfer function
\[
|H_1(f)|^2=2\sin^2(\pi f_x),\qquad
|H_2(f)|^2=2\sin^2(\pi f_y),\qquad
|H_3(f)|^2=2\sin^2\bigl(\pi (f_x+f_y)\bigr).
\]
Each of the three difference planes $d_m$, $m=1,2,3$, is winsorized on its own scale, with $\hat\sigma_m=1.4826\,\operatorname{med}\bigl|d_m-\operatorname{med}d_m\bigr|$ and $d_m\leftarrow\operatorname{clip}(d_m,\pm 4\hat\sigma_m)$, the median and the deviation being taken over all pixels of that plane. Its periodogram $P_d$ is a Welch average over $L\times L$ patches on a stride of $L/2$, each patch having its own mean removed, multiplied by a separable Hann window $w$, and normalized by $L^{2}\langle w^{2}\rangle$. We use $L=128$. If either side of the plane is shorter than 128, $L$ becomes the largest even number not exceeding the shorter side, with a floor of 16, and the stride becomes half of that, with a floor of 8. Patches that would extend past the last row or column are dropped. If no patch fits, which happens when the shorter side is below 16, the four structural statistics are set to zero.

Since $\mathbb{E}P_d=|H_d|^2S$, the three periodograms are combined by least squares, $\hat S=\arg\min_S\sum_d\bigl(P_d-|H_d|^2S\bigr)^2$, with a ridge term in the denominator,
\[
\hat S(f)=\frac{\sum_{d=1}^{3}|H_d(f)|^2\,P_d(f)}{\sum_{d=1}^{3}|H_d(f)|^4+10^{-6}},
\qquad \hat S(0)=0.
\]
Each direction therefore contributes with weight $|H_d|^2$, which vanishes on the line of frequencies its own difference does not transmit: $f_x=0$ for $d_1$, $f_y=0$ for $d_2$, and $f_x+f_y=0$ for $d_3$. No single difference determines $S$ on its own blind line, and any two of the three have blind lines that meet only at the origin, where $\hat S$ is set to zero.

\paragraph{Structural statistics (10--13).} All four are computed on the annulus $\mathcal{M}=\{f:0.02<\|f\|\le 0.5\}$, which excludes the lowest frequencies, where image content dominates, and the corners of the square, which lie beyond Nyquist. Write $S=\hat S|_{\mathcal{M}}$, $|\mathcal{M}|$ for the number of bins in the annulus, and $S_{(1)}\le\dots\le S_{(|\mathcal{M}|)}$ for the sorted values of $S$. The first two statistics measure how concentrated the spectrum is,
\[
\texttt{peak\_prom}=\log\frac{\max S}{\operatorname{med} S},
\qquad
\texttt{line\_frac}=\log\frac{\sum_{i>|\mathcal{M}|-\kappa}S_{(i)}}{\sum_i S_{(i)}}, \qquad \kappa=\max\bigl(\lfloor 0.005\,|\mathcal{M}|\rfloor,1\bigr),
\]
the height of the largest bin above the bulk of the spectrum, measured against the median, and the share of the energy carried by the strongest half percent of the bins. Neither involves detecting a peak or a line, and neither refers to a frequency, so a periodic corruption keeps its value when its frequency changes.

The remaining two measure how the energy is distributed in direction. Let $f^\star=\arg\max_{\mathcal{M}} S$ and let $\mathcal{R}=\bigl\{f\in\mathcal{M}: 0.85\,\|f^\star\|<\|f\|<1.15\,\|f^\star\|\bigr\}$ be the annulus at the radius of the maximum. Then
\[
\texttt{ang\_conc}=\log\frac{\max_{\mathcal{R}} S}{\operatorname{mean}_{\mathcal{R}} S},
\qquad
\texttt{anisotropy}=\frac{\lambda_1-\lambda_2}{\lambda_1+\lambda_2}
=\frac{\sqrt{(M_{xx}-M_{yy})^2+4M_{xy}^2}}{M_{xx}+M_{yy}},
\]
where $\texttt{ang\_conc}$ is set to zero if $|\mathcal{R}|\le 8$, and $\lambda_1\ge\lambda_2$ are the eigenvalues of the second-moment matrix of the spectrum, $M_{pq}=\sum_{\mathcal{M}} S\,f_p f_q \big/ \sum_{\mathcal{M}} S$ for $p,q\in\{x,y\}$. No centring appears because the spectrum of a real field satisfies $S(-f)=S(f)$, so its first moment vanishes and the second moment about the origin is the covariance. A ring and a pair of isolated peaks have the same radial profile, and \texttt{ang\_conc} separates them; \texttt{anisotropy} reports the strength of a preferred direction without reporting which direction it is. That is a deliberate omission and it has a cost: horizontal and vertical banding differ in no coordinate by as much as its noise floor (Figure~\ref{fig:sensitivity}), which is why routing sends vertical banding to a specialist trained on horizontal stripes (Section~\ref{sec:families}).
Both depend on where in the spectrum the structure lies, \texttt{ang\_conc} through the number of bins in $\mathcal{R}$ and \texttt{anisotropy} through the weighting by $f_pf_q$, so both read a low-frequency line as weaker than a high-frequency one of the same contrast.

All four are ratios and are therefore unchanged when the noise is scaled: multiplying the image noise by a constant multiplies $\hat S$ by its square and leaves every one of them at its previous value. None of the four is zero on white noise, however: the residual fluctuation of a Welch average sets a floor, on a $256\times256$ plane $1.33$, $-4.25$, $1.24$ and $0.02$ respectively, against $9.70$, $-0.16$, $4.80$ and $0.49$ once a sinusoid is added. What the statistics report is the excursion above that floor.

\paragraph{Robustness choices.} Several of the steps above guard against outliers and against leakage between frequencies. Winsorization at four median absolute deviations, applied before every periodogram, prevents impulsive pixels from spreading energy across the whole spectrum; removing the mean of each patch before the window stops a difference plane with a nonzero mean from leaking its offset in the same way; and the median, rather than the mean, serves as the reference level in \texttt{peak\_prom} so that the peak itself does not raise the level it is measured against. The Welch average reduces the variance of the periodogram at the cost of resolution, and the ridge of $10^{-6}$ bounds the estimate near the blind lines. The three concentration statistics are reported as logarithms because their raw ratios span several orders of magnitude. A bin of the intensity sweep with fewer than eight samples is replaced by the global variance, and any coordinate that is not finite is replaced by zero.

\paragraph{Cost.} The descriptor is computed on the CPU, with no GPU and no trained component. On a $256\times256$ RGB image it takes 23.7~ms on one core of an Apple M2~Max, the median over 50 photographs of the benchmark.

\subsection{Sensitivity to the corruption parameters}
\label{app:sensitivity}
\begin{figure}[H]
\centering
\includegraphics[width=1\linewidth]{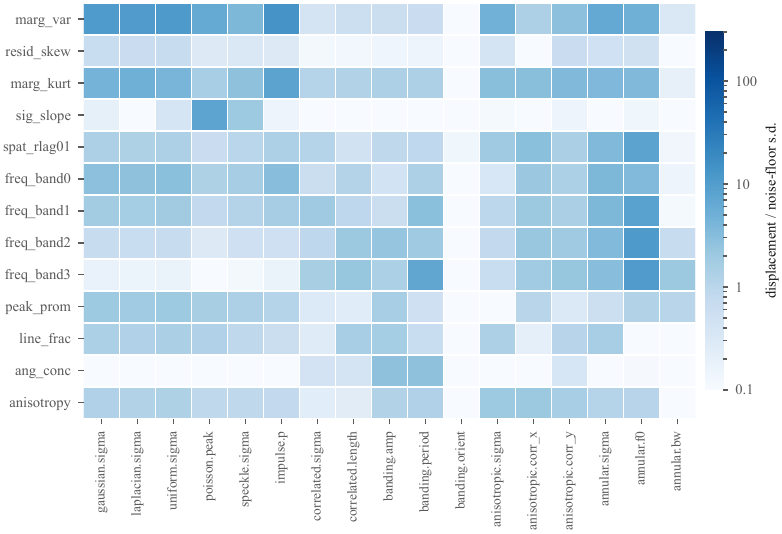}
\caption{Sensitivity of the \Dxiii statistics to the corruption parameters. Each column varies one parameter of one corruption family while the others are held fixed, and each row is one statistic. Each cell reports how far the statistic moves across the sweep, divided by its standard deviation across photographs and noise realizations. Values above one indicate that the statistic moves more with the parameter than with the photograph and the noise realization; darker cells indicate larger values. The statistics are nearly insensitive to the orientation of banding (see Appendix~\ref{app:statistics}).}
\label{fig:sensitivity}
\end{figure}

To check that each statistic responds to the property it is meant to measure, we varied one parameter of one family at a time, holding the others at the family's reference values, for 17 parameters of the ten families. Each parameter took 13 values over the range given in Appendix~\ref{app:corruptions}, and each value was applied to 300 photographs, 30 from each content domain, with independent noise realizations. For each statistic and parameter we measured the displacement of the statistic across the sweep, the range between its largest and smallest value over the 13 parameter values, each averaged over the 300 photographs. We divided this by the standard deviation of the statistic over the 300 photographs and five independent noise realizations at the family's reference values, which is its variation at a fixed corruption (Figure~\ref{fig:sensitivity}). The ratio exceeded 1 for 16 of the 17 parameters.

The variance of the residual responded to every amplitude parameter except those of correlated and banding noise, and the structural statistics to the parameters of the spatially structured families. The response was weakest for the orientation of banding and the bandwidth of annular noise, and it was low for the amplitude and the correlation length of correlated noise, whose energy lies mostly at the low frequencies that the residual attenuates. The structural statistics measure how concentrated and how directional the spectrum is, not in which direction, so they do not distinguish horizontal from vertical stripes. The classical pool is unaffected, since its notch filter removes stripes of either orientation, but the specialist pool is not (Section~\ref{sec:families}).

\section{Pools of denoisers}
\label{app:pools}

Table~\ref{tab:pools} lists every candidate of the four pools of Section~\ref{sec:pools}, with the corruption family each is meant for, the implementation or the publication it comes from, and the settings at which it is run. The notes below the table give the library versions, the checkpoints used for the pretrained networks, and the composition of the hybrid pool, whose candidates are shared with the other pools apart from the four listed separately. A candidate is run at these settings on every image, since no candidate is retrained or tuned during routing.

\begin{table}[t]
\centering
\caption{Candidates in each pool.}
\label{tab:pools}

\fontsize{8}{9}\selectfont
\setlength{\tabcolsep}{2.2pt}
\renewcommand{\arraystretch}{0.93}

\begin{tabularx}{\textwidth}{
    @{}
    L{1.15cm}
    L{3.55cm}
    L{2.35cm}
    L{2.25cm}
    Y
    @{}
}
\toprule
Pool & Candidate & Target family & Source & Settings \\
\midrule

Classical
& \texttt{gauss\_s0.6}
& broadband additive
& SciPy
& Gaussian blur, $\sigma=0.6$ px \\

& \texttt{gauss\_s1.2}
& broadband additive
& SciPy
& Gaussian blur, $\sigma=1.2$ px \\

& \texttt{gauss\_s2.4}
& broadband additive
& SciPy
& Gaussian blur, $\sigma=2.4$ px \\

& \texttt{median\_3}
& impulse
& SciPy
& median filter, $3\times3$ \\

& \texttt{median\_5}
& impulse
& SciPy
& median filter, $5\times5$ \\

& \texttt{tv\_0.05}
& none
& scikit-image, \citet{chambolle2004algorithm}
& TV-Chambolle,  weight $0.05$  \\

& \texttt{tv\_0.15}
& none
& scikit-image, \citet{chambolle2004algorithm}
& TV-Chambolle, weight $0.15$  \\

& \texttt{bilateral}
& none
& scikit-image, \citet{tomasi1998bilateral}
& $\sigma_{\mathrm{col}}=0.10$, $\sigma_{\mathrm{sp}}=3$ \\

& \texttt{wiener\_5}
& none
& SciPy
& local Wiener, $5\times5$ \\

& \texttt{notch}
& periodic
& ours
& FFT bins $3.5$ sd above a $7\times7$ median background,
  $f>0.06$ \\

\midrule

Specialist
& \texttt{M\_gaussian\_15}
& Gaussian
& ours
& $\sigma=15$ \\

& \texttt{M\_gaussian\_25}
& Gaussian
& ours
& $\sigma=25$ \\

& \texttt{M\_gaussian\_50}
& Gaussian
& ours
& $\sigma=50$ \\

& \texttt{M\_laplacian\_25}
& Laplacian
& ours
& $\sigma=25$ \\

& \texttt{M\_uniform\_25}
& uniform
& ours
& $\sigma=25$ \\

& \texttt{M\_poisson\_30}
& Poisson
& ours
& peak $=30$ \\

& \texttt{M\_poisson\_10}
& Poisson
& ours
& peak $=10$ \\

& \texttt{M\_speckle\_10}
& speckle
& ours
& $\sigma=0.10$ \\

& \texttt{M\_speckle\_20}
& speckle
& ours
& $\sigma=0.20$ \\

& \texttt{M\_sp\_5pct}
& impulse
& ours
& $p=0.05$ \\

& \texttt{M\_corr\_gauss}
& correlated
& ours
& $\sigma=25$, length $=3$ \\

& \texttt{M\_banding}
& banding
& ours
& amp $=20$, period $=10$ \\

& \texttt{M\_aniso\_h}
& anisotropic
& ours
& $\sigma=25$, $(8,1)$ \\

& \texttt{M\_annular\_mid}
& annular
& ours
& $\sigma=25$, $f_0=0.25$ \\

\midrule

Pretrained
& \texttt{S\_DRUNet\_s15}
& Gaussian
& \citet{zhang2021plug}
& noise-level input $15$ \\

& \texttt{S\_DRUNet\_s25}
& Gaussian
& \citet{zhang2021plug}
& noise-level input $25$ \\

& \texttt{S\_DRUNet\_s50}
& Gaussian
& \citet{zhang2021plug}
& noise-level input $50$ \\

& \texttt{S\_SwinIR}
& Gaussian
& \citet{liang2021swinir}
& color, $\sigma=15$ \\

& \texttt{S\_Restormer\_gau}
& Gaussian
& \citet{zamir2022restormer}
& blind over $\sigma$, color \\

& \texttt{S\_Restormer\_real}
& real camera noise (SIDD)
& \citet{zamir2022restormer}
& real-noise checkpoint \\

& \texttt{S\_SCUNet\_real}
& real-world degradations, synthesized
& \citet{zhang2023practical}
& real-noise checkpoint \\

& \texttt{S\_PromptIR}
& blind, all-in-one
& \citet{potlapalli2023promptir}
& authors' checkpoint: denoising, deraining, dehazing \\

\midrule

Hybrid
& \texttt{S\_SwinIR\_s25}
& Gaussian
& \citet{liang2021swinir}
& color, $\sigma=25$ \\

& \texttt{S\_SwinIR\_s50}
& Gaussian
& \citet{liang2021swinir}
& color, $\sigma=50$ \\

& \texttt{S\_NAFNet\_SIDD\_w32}
& real camera noise
& \citet{chen2022simple}
& width $32$, SIDD \\

& \texttt{adaptive\_median\_w7}
& impulse
& \citet{gonzalez2018digital}
& max window $7$ \\

\bottomrule
\end{tabularx}

\vspace{2pt}

\begin{minipage}{\textwidth}
\scriptsize
\textit{Implementation details.}
Classical operators use scikit-image 0.26.0
\citep{scikit-image} and SciPy 1.17.1
\citep{2020SciPy-NMeth}; learned models use PyTorch 2.12.1
\citep{paszke2019pytorch}.
Classical operators are applied to each RGB channel separately; the networks take the RGB image as input.
The eight pretrained networks are loaded through DeepInverse
\citep{tachella2025deepinverse} (deepinv 0.4.1), using its default
released checkpoints:
\texttt{drunet\_deepinv\_color\_finetune\_22k},
\texttt{005\_colorDN\_DFWB\_s128w8\_SwinIR-M\_noise\{15,25,50\}},
\texttt{gaussian\_color\_denoising\_blind},
\texttt{real\_denoising},
\texttt{scunet\_color\_real\_psnr}, and
\texttt{promptir.ckpt}.

\textit{Hybrid pool.}
It reuses unchanged the eight pretrained networks, with
\texttt{S\_SwinIR} renamed \texttt{S\_SwinIR\_s15}, plus
\texttt{median\_3}, \texttt{median\_5},
\texttt{tv\_0.05}, \texttt{tv\_0.15}, and \texttt{notch};
the four additional candidates are listed above.
\end{minipage}

\end{table}

\paragraph{Specialist networks.} The fourteen specialists share one architecture and one training procedure, and differ only in the corruption applied during training. Each is a U-Net \citep{ronneberger2015unet} with three resolution levels and 32 channels at the first level, doubled at each downsampling, for 468k parameters in total. It differs from the original in four ways: three levels instead of five; padded $3\times3$ convolutions, so the output has the size of the input and the skip connections need no cropping; batch normalization after every convolution; and a $1\times1$ convolution to three channels followed by a sigmoid in place of the two-class output. The network predicts the clean image directly, on the unit intensity scale. Each specialist was trained on $128\times128$ patches cropped from the 800 training photographs of DIV2K \citep{agustsson2017ntire}, corrupted with a new realization of its calibration condition at every step by the benchmark's own generator from a disjoint seed stream, with random flips and multiples of ninety degrees, the $\ell_2$ loss and Adam \citep{kingma2014adam} at learning rate $2\times10^{-4}$ with cosine decay for 30{,}000 iterations and batch size 32, in mixed precision. Validation PSNR was measured every 2{,}000 iterations on fixed $256\times256$ crops of the 100 DIV2K validation photographs with a fixed noise realization, and the best checkpoint was kept. DIV2K shares no photograph and no content domain with the benchmark. Training one specialist took about twelve minutes on one NVIDIA A100 80GB, and under three hours for all fourteen.

\begingroup
\section{All candidates in one pool}
\label{app:merged}

The four pools of Section~\ref{sec:pools} share some candidates, since the hybrid pool reuses the eight pretrained networks and five of the classical operators. Together they contain 36 distinct candidates (Table~\ref{tab:pools}). ``Without classical'' removes the 10 classical operators and the classical operator added in the hybrid pool, and ``without pretrained'' removes the 8 pretrained networks and the 3 networks added in the hybrid pool. We routed over all 36 with the procedure of Section~\ref{sec:routing}, and over the same set with the candidates of one pool removed, on the 26 withheld conditions of the discrete benchmark (Table~\ref{tab:merged}).

Routing over all 36 candidates gained 1.20~dB over the best overall denoiser, \texttt{S\_SCUNet\_real}, and reached the highest routed PSNR of any pool. Removing the classical operators left the routed PSNR unchanged. Removing the specialist networks or the pretrained networks lowered it by 0.22 and 0.19~dB. The oracle remained 1.9~dB above routing.

\begin{table}[h]

\centering
\caption{Routing over all 36 candidates, over the same set with one pool's candidates removed, and over each pool separately, on the 26 withheld conditions. $K$ is the number of candidates. \emph{Fixed}, \emph{Routed}, and \emph{Oracle} are mean PSNR (dB) of the best overall denoiser, of routing, and of the per-image best candidate. \emph{Gain} is routed minus fixed. The last column is the routed PSNR over all 36 candidates minus that of the row. Intervals are 95\% paired intervals, and bold marks an interval that excludes zero.}
\label{tab:merged}
\resizebox{\linewidth}{!}{%
\begin{tabular}{lrlccccc}
\toprule
Candidates & $K$ & Best overall & Fixed & Routed & Oracle & Gain & All 36 $-$ row \\
\midrule
All 36             & 36 & \texttt{S\_SCUNet\_real}  & 26.48 & 27.68 & 29.54 & $\mathbf{+1.20}$ {\scriptsize[0.98, 1.36]} & -- \\
Without classical  & 25 & \texttt{S\_SCUNet\_real}  & 26.48 & 27.68 & 29.05 & $\mathbf{+1.20}$ {\scriptsize[0.98, 1.36]} & $+0.00$ {\scriptsize[0.00, 0.00]} \\
Without specialist & 22 & \texttt{S\_SCUNet\_real}  & 26.48 & 27.46 & 28.77 & $\mathbf{+0.99}$ {\scriptsize[0.71, 1.16]} & $\mathbf{+0.22}$ {\scriptsize[0.10, 0.33]} \\
Without pretrained & 25 & \texttt{M\_laplacian\_25} & 25.91 & 27.49 & 28.96 & $\mathbf{+1.58}$ {\scriptsize[1.48, 1.67]} & $\mathbf{+0.19}$ {\scriptsize[0.11, 0.26]} \\
\midrule
Classical          & 10 & \texttt{tv\_0.15}         & 24.36 & 25.73 & 26.41 & $\mathbf{+1.36}$ {\scriptsize[1.21, 1.49]} & $\mathbf{+1.95}$ {\scriptsize[1.63, 2.26]} \\
Specialist         & 14 & \texttt{M\_laplacian\_25} & 25.91 & 27.49 & 28.46 & $\mathbf{+1.58}$ {\scriptsize[1.48, 1.67]} & $\mathbf{+0.19}$ {\scriptsize[0.11, 0.26]} \\
Pretrained         &  8 & \texttt{S\_SCUNet\_real}  & 26.48 & 26.62 & 27.40 & $+0.14$ {\scriptsize[$-0.01$, 0.27]} & $\mathbf{+1.06}$ {\scriptsize[0.95, 1.15]} \\
Hybrid             & 17 & \texttt{S\_SCUNet\_real}  & 26.48 & 27.46 & 28.76 & $\mathbf{+0.99}$ {\scriptsize[0.71, 1.16]} & $\mathbf{+0.22}$ {\scriptsize[0.10, 0.33]} \\
\bottomrule
\end{tabular}}
\end{table}
\endgroup

\section{Benchmarks}\label{app:benchmarks}

We describe the discrete and parametric evaluation suites, including their calibration conditions, held-out corruptions, content splits, and evaluation protocols.

The discrete suite tests whether routing transfers to unseen corruption conditions and compositions, including mixtures. The parametric suite isolates a different question: within each primitive corruption family, how does routing behave as one physical parameter moves within or beyond the calibrated range?

\paragraph{Photographs.} The ten content domains are the ImageNette classes: tench, English springer, cassette player, chain saw, church, French horn, garbage truck, gas pump, golf ball, and parachute \citep{howard2019imagenette}\rev{, a subset of ImageNet}. Each photograph was resized to a short side of 320 pixels and center-cropped to $256\times256$ RGB pixels. Each benchmark used thirty photographs per domain, and the photographs of the parametric benchmark were not used in the discrete benchmark.

\paragraph{Discrete benchmark.} The discrete benchmark has 40 conditions, of which 14 form the calibration set (Table~\ref{tab:calibration}). The calibration set contains at least one condition from each of the ten families of Appendix~\ref{app:corruptions} and no mixture. The other 26 conditions are withheld: twelve single conditions that change the severity or the orientation of a calibrated family, and the fourteen mixtures. None of the withheld conditions introduces a family absent from calibration. Each network of the specialist pool was trained on one calibration condition (Appendix~\ref{app:pools}).

\begin{table}[t]
\centering
\small
\setlength{\tabcolsep}{5pt}
\caption{The 14 calibration conditions of the discrete benchmark. The correlation length $\ell$ is given in pixels, and $\ell_x$ and $\ell_y$ are the correlation lengths along the horizontal and vertical axes (Appendix~\ref{app:corruptions}).}\label{tab:calibration}
\begin{tabular}{llll}
\toprule
\# & Condition & Family & Parameters \\
\midrule
1  & \texttt{gaussian\_15}  & Gaussian     & $\sigma=15$ \\
2  & \texttt{gaussian\_25}  & Gaussian     & $\sigma=25$ \\
3  & \texttt{gaussian\_50}  & Gaussian     & $\sigma=50$ \\
4  & \texttt{laplacian\_25} & Laplacian    & $\sigma=25$ \\
5  & \texttt{uniform\_25}   & Uniform      & $\sigma=25$ \\
6  & \texttt{sp\_5pct}      & Impulse      & $p=0.05$ \\
\addlinespace
7  & \texttt{poisson\_30}   & Poisson      & peak $=30$ \\
8  & \texttt{poisson\_10}   & Poisson      & peak $=10$ \\
9  & \texttt{speckle\_10}   & Speckle      & $\sigma=0.10$ \\
10 & \texttt{speckle\_20}   & Speckle      & $\sigma=0.20$ \\
\addlinespace
11 & \texttt{corr\_gauss}   & Correlated   & $\sigma=25$, $\ell=3$ \\
12 & \texttt{aniso\_h}      & Anisotropic  & $\sigma=25$, $\ell_x=8$, $\ell_y=1$ \\
13 & \texttt{banding}       & Banding      & amplitude $20$, period $10$, horizontal \\
14 & \texttt{annular\_mid}  & Annular      & $\sigma=25$, $f_0=0.25$, $\mathrm{bw}=0.04$ \\
\bottomrule
\end{tabular}
\end{table}

\paragraph{Parametric benchmark.} For each family, the six severity levels were drawn from equal strata of the parameter's range. The parametric benchmark uses the leave-one-domain-out protocol of Section~\ref{sec:protocol}, with all thirty photographs of each domain, and is calibrated separately for each of the splits S1 to S3. Table~\ref{tab:parametric-levels} lists the parameter ranges and levels.

\paragraph{Confidence intervals.} Each photograph was corrupted once under each condition, with an independent noise realization, so a condition is evaluated over 300 realizations; no condition is repeated under a second seed. Each bootstrap replicate draws ten content domains with replacement and, within each drawn domain, draws its photographs with replacement. Conditions are not resampled, since they define the benchmark rather than sample from a population. All methods are evaluated on the same resampled images, so their differences remain paired in every resample. Intervals are percentile intervals over $5{,}000$ replicates.

\begin{table}[t]
\centering
\caption{Parameter levels used in the parametric benchmark.}
\label{tab:parametric-levels}

\small
\setlength{\tabcolsep}{2.2pt}
\renewcommand{\arraystretch}{1.02}

\begin{tabular*}{\textwidth}{
    @{\extracolsep{\fill}}
    p{1.65cm}
    p{3.15cm}
    p{1.55cm}
    rrrrrr
    @{}
}
\toprule
Family & Parameter & Range & 1 & 2 & 3 & 4 & 5 & 6 \\
\midrule

Gaussian
& $\sigma$
& 10--50
& 10.0 & 22.7 & 29.9 & 35.2 & 36.7 & 45.5 \\

Laplacian
& $\sigma$
& 10--50
& 15.5 & 22.0 & 25.2 & 31.6 & 41.4 & 48.1 \\

Uniform
& $\sigma$
& 10--50
& 13.2 & 21.5 & 23.6 & 32.8 & 41.7 & 46.2 \\

Poisson
& peak count
& 8--60
& 58.3 & 49.0 & 42.0 & 33.2 & 23.4 & 13.4 \\

Speckle
& $\sigma$
& 0.04--0.25
& 0.060 & 0.079 & 0.112 & 0.157 & 0.189 & 0.220 \\

Impulse
& $p$
& 0.01--0.12
& 0.018 & 0.030 & 0.055 & 0.071 & 0.091 & 0.119 \\

Correlated
& correlation length
& 1.5--8
& 2.19 & 3.00 & 4.06 & 5.13 & 6.83 & 7.51 \\

Banding
& period
& 5--30
& 7.89 & 9.23 & 15.8 & 20.5 & 25.3 & 29.2 \\

Anisotropic
& horizontal correlation length
& 1--16
& 2.41 & 3.57 & 6.97 & 10.1 & 11.9 & 14.9 \\

Annular
& center frequency $f_0$
& 0.08--0.42
& 0.407 & 0.362 & 0.260 & 0.226 & 0.156 & 0.092 \\

\bottomrule
\end{tabular*}

\vspace{2pt}

\parbox{\textwidth}{\footnotesize
\textit{Notes.}
Parameters held fixed: $\sigma=25$ for correlated, anisotropic,
and annular noise; vertical correlation length 1 for anisotropic
noise; bandwidth 0.04 for annular noise; amplitude 20 and orientation 0 for banding.
The six levels are sampled from equal strata of each stated range.
The level index increases with severity rather than parameter value: for Poisson and annular noise, severity increases as the parameter decreases, so their values decrease from left to right.
}

\end{table}

\section{No-reference selectors}
\label{app:selectors}

Each selector ran every candidate of the pool on a test image, scored each output without the clean image, and kept the output with the best score. The six perceptual scores were computed with IQA-PyTorch \citep{pyiqa}, version 0.1.16, each through \texttt{pyiqa.create\_metric} with the authors' default checkpoints; the direction is read from the metric itself rather than assumed, so the lowest score is kept for NIQE and BRISQUE and the highest for TOPIQ, MUSIQ, CLIP-IQA+ and ARNIQA. All selectors were evaluated on the same test images as routing, and the PSNR of each selected output was taken from the same recorded outputs.

\paragraph{Monte-Carlo SURE.} For an image $y$ corrupted by Gaussian noise of variance $\sigma^2$, and a denoiser $f$, Stein's unbiased risk estimate of the mean squared error is
\begin{equation}
\mathrm{SURE}(f, y) = \frac{1}{N}\,\|y - f(y)\|^2 - \sigma^2 + \frac{2\sigma^2}{N}\,\operatorname{div} f(y),
\end{equation}
where $N$ is the number of pixels. Following \citet{ramani2008monte}, we estimated the divergence with one random probe $b$ of independent standard normal entries,
\begin{equation}
\operatorname{div} f(y) \approx \frac{1}{\epsilon}\, b^{\top}\bigl(f(y + \epsilon b) - f(y)\bigr),
\end{equation}
with $\epsilon = 1$ in intensity units on the $[0, 255]$ scale so each candidate requires two denoising calls. The probe seed is fixed and depends deterministically on the candidate, so the ranking is reproducible from a single probe. The divergence is estimated for the clipped map, which is the map whose risk is being ranked. The variance $\sigma^2$ was estimated from the corrupted image, as the square of the median absolute deviation of the finest-scale diagonal Haar wavelet coefficients divided by 0.6745 \citep{donoho1994ideal}; one estimate was computed per test image and shared by all candidates. We checked the implementation in three steps. The Monte-Carlo divergence reproduced the analytic trace of a linear shift-invariant filter within $0.2\%$. On an unclipped Gaussian perturbation with known $\sigma$, where SURE is exactly unbiased, it reproduced the true mean squared error within $0.04\%$. Clipping the output to the intensity range, which every candidate does, biases the estimate downward by $11\%$, and on the trained candidates at \texttt{gaussian\_35} it sits 17 to 21\% below the true error while still ranking them in the order the true error does; the selector uses only that ranking.

Routing selected the oracle candidate, the one with the highest PSNR on that image, in 56, 46, 49, and 40\% of images for the classical, specialist, pretrained, and hybrid pools (Table~\ref{tab:D13_arms}), and the selectors in 2 to 39\% (Table~\ref{tab:selectors}). 
 
\begin{table}[t]
\centering
\caption{Routing with \Dxiii on the withheld conditions of the discrete benchmark. Gain in PSNR (dB) over the best overall denoiser, with 95\% paired-bootstrap intervals, and the oracle gap for the same pool. \emph{Recovered} is the fraction of the oracle gap that routing recovers, and \emph{optimal} the fraction of images on which routing selects the candidate with the highest PSNR.}
\label{tab:D13_arms}
\begin{tabular}{lcccrr}
\toprule
Pool & $K$ & \Dxiii routing & Oracle & Recovered & Optimal \\
\midrule
Classical       & 10 & $+1.363$ $[+1.212, +1.492]$ & $+2.050$ & $0.665$ & $0.563$ \\
Specialist  & 14 & $+1.577$ $[+1.479, +1.669]$ & $+2.542$ & $0.621$ & $0.456$ \\
Pretrained &  8 & $+0.142$ $[-0.009, +0.265]$ & $+0.928$ & $0.153$ & $0.485$ \\
Hybrid          & 17 & $+0.986$ $[+0.710, +1.161]$ & $+2.282$ & $0.432$ & $0.404$ \\
\bottomrule
\end{tabular}
\end{table}

\begin{table}[h]
\centering
\caption{Gain in PSNR (dB) over the best overall denoiser on the withheld conditions of the discrete benchmark. Calls denote number of denoising calls per image for a pool of $K$ candidates.}
\label{tab:selectors}
\begin{tabular}{lccccc}
\toprule
Selector & Calls & Classical & Specialist & Pretrained & Hybrid \\
\midrule
Routing (\Dxiii) & 1    & $+1.36$ & $+1.58$ & $+0.14$ & $+0.99$ \\
NIQE             & $K$  & $+0.86$ & $+1.17$ & $-1.25$ & $-0.92$ \\
BRISQUE          & $K$  & $+0.05$ & $+0.90$ & $-1.43$ & $-1.33$ \\
MUSIQ            & $K$  & $-0.87$ & $+0.41$ & $-0.35$ & $-0.05$ \\
CLIP-IQA+        & $K$  & $+0.02$ & $+0.29$ & $-0.26$ & $+0.38$ \\
TOPIQ            & $K$  & $-0.45$ & $+0.16$ & $-0.33$ & $+0.20$ \\
ARNIQA           & $K$  & $-0.47$ & $+0.50$ & $-0.33$ & $+0.17$ \\
MC-SURE          & $2K$ & $-0.97$ & $-1.23$ & $-2.53$ & $-3.54$ \\
\midrule
Oracle           & ---  & $+2.05$ & $+2.54$ & $+0.93$ & $+2.28$ \\
\bottomrule
\end{tabular}
\end{table}

\paragraph{SURE by condition.} Table~\ref{tab:sure} gives, for each withheld condition, the rank correlation between the SURE estimate and the true mean squared error across the candidates of the specialist pool, averaged over test images. The mean over the 26 conditions was 0.050; the correlation was positive for 13 of them and zero to two decimals for two more. It was close to one on Gaussian noise and close to minus one on impulse noise, and it was negative on the anisotropic and correlated conditions, whose noise is not independent across pixels, and on annular noise at low center frequency, though not at high center frequency. Nothing in the estimate itself indicates which of these cases applies to a given image.

\begin{table}[t]
\centering
\caption{Rank correlation between the SURE estimate and the true mean squared error for each of the 26 withheld conditions. For each image, the Spearman correlation is computed across the candidates of the specialist pool; each entry is its mean over the 300 images of the condition.}
\label{tab:sure}
\begin{tabular}{@{}llr@{}}
\toprule
Condition & Family & Correlation \\
\midrule
Gaussian, $\sigma = 35$                        & additive    & $+0.89$ \\
30\% anisotropic + 70\% Gaussian $\sigma=25$   & mixture     & $+0.88$ \\
60\% Gaussian $\sigma=25$ + 40\% Poisson 30    & mixture     & $+0.87$ \\
30\% correlated + 70\% Gaussian $\sigma=25$    & mixture     & $+0.86$ \\
Annular, $f_0 = 0.40$                          & annular     & $+0.84$ \\
30\% Gaussian $\sigma=25$ + 70\% Poisson 30    & mixture     & $+0.77$ \\
30\% banding + 70\% Poisson 30                 & mixture     & $+0.62$ \\
Speckle, $\sigma = 0.15$                       & signal-dep. & $+0.59$ \\
Poisson, peak 20                               & signal-dep. & $+0.58$ \\
60\% banding + 40\% Poisson 30                 & mixture     & $+0.33$ \\
Banding, vertical                              & banding     & $+0.18$ \\
Banding, period 14                             & banding     & $+0.16$ \\
60\% anisotropic + 40\% Gaussian $\sigma=25$   & mixture     & $+0.12$ \\
60\% correlated + 40\% Gaussian $\sigma=25$    & mixture     & $+0.00$ \\
30\% correlated + 70\% banding                 & mixture     & $-0.00$ \\
60\% correlated + 40\% banding                 & mixture     & $-0.12$ \\
Correlated, length 8                           & correlated  & $-0.31$ \\
60\% Gaussian $\sigma=25$ + 40\% impulse 5\%   & mixture     & $-0.34$ \\
Correlated, length 5                           & correlated  & $-0.51$ \\
Anisotropic, horizontal, length 16             & anisotropic & $-0.56$ \\
Anisotropic, vertical                          & anisotropic & $-0.61$ \\
Annular, $f_0 = 0.10$                          & annular     & $-0.71$ \\
60\% annular + 40\% impulse 5\%                & mixture     & $-0.72$ \\
30\% Gaussian $\sigma=25$ + 70\% impulse 5\%   & mixture     & $-0.78$ \\
30\% annular + 70\% impulse 5\%                & mixture     & $-0.83$ \\
Impulse, $p = 0.08$                            & impulse     & $-0.93$ \\
\bottomrule
\end{tabular}
\end{table}

\section{Transfer across noise severities}
\label{app:parametric}

\begin{figure}[h]
    \centering
    \includegraphics[width=0.85\linewidth]{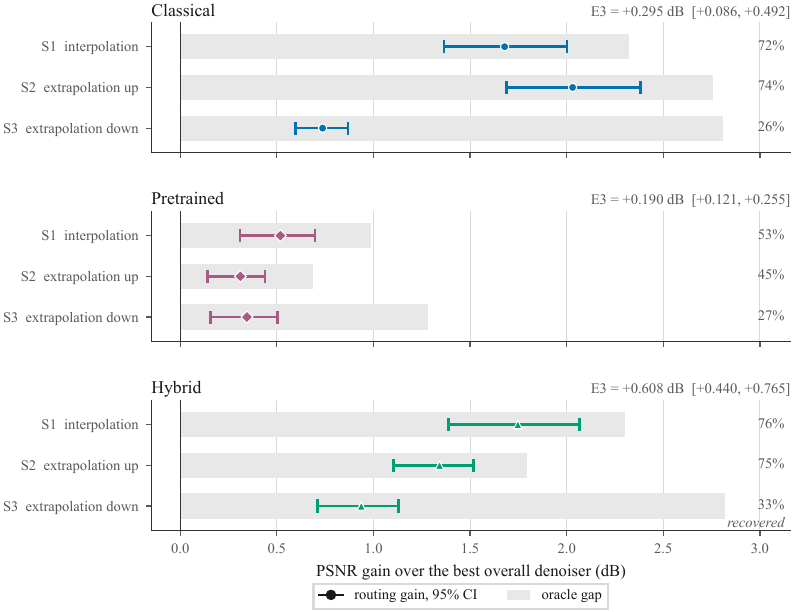}
    \caption{\textbf{Transfer across noise severities.} Gain in PSNR over the best overall denoiser on the parametric benchmark, at severities between the calibrated ones (S1), above them (S2), and below them (S3). Markers and whiskers show routing gain and its 95\% interval, gray bars show the oracle gap, and percentages give the fraction of it recovered. The specialist pool is not evaluated on this benchmark.} 
    \label{fig:parametric}
\end{figure}

Figure~\ref{fig:parametric} gives the gains of Section~\ref{sec:parametric} for each split. The loss of gain outside the calibrated range, measured as the gain on S1 minus the mean gain on S2 and S3, was 0.30~dB for the classical pool and 0.61~dB for the hybrid pool, against 0.19~dB for the pretrained pool, with 95\% intervals lying entirely above zero. In the classical and hybrid pools, strong noise dominates the statistics the descriptor measures and is estimated accurately, whereas mild noise changes them little, so severities below the calibrated range crowd together near the mildest calibrated reference.

The specialist pool is omitted because each of its networks was trained on one calibration condition of the discrete benchmark. The correspondence between candidates and calibration conditions, which this pool was built to test, does not hold for the calibration sets of the parametric benchmark unless the networks are retrained for each split.

\section{Gains by corruption family}
\label{app:families}

\begin{figure}[t]
\centering
\includegraphics[width=.85\linewidth]{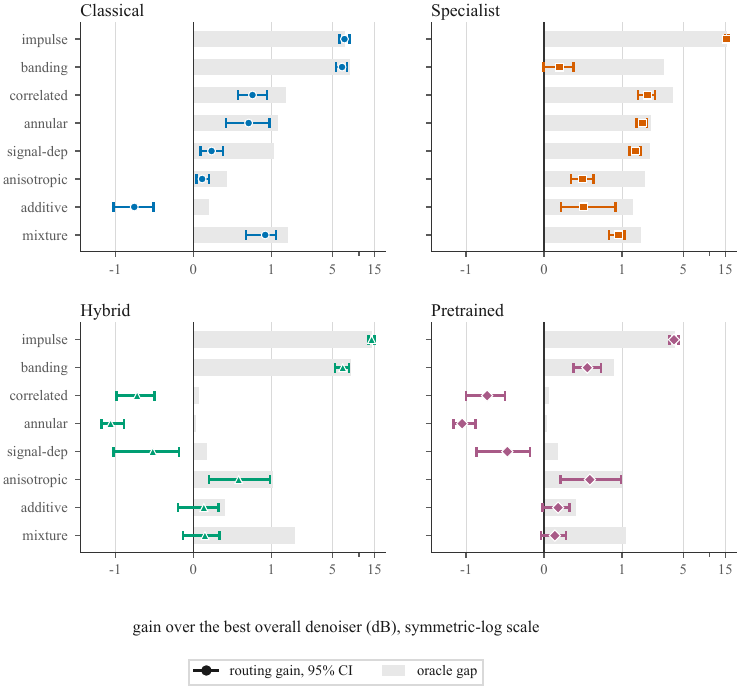}
\caption{Gain of routing over the best overall denoiser, by corruption family and pool, on the withheld conditions of the discrete benchmark. Points and whiskers show the gain and its 95\% interval; gray bars show the oracle gap. The horizontal axis is linear between $-1$ and 1~dB and logarithmic beyond, so that the gains on impulse noise and on the other families can be read on one axis.}
\label{fig:families}
\end{figure}

Figure~\ref{fig:families} gives the gains of Table~\ref{tab:families} with their 95\% intervals, computed by the paired bootstrap of Section~\ref{sec:protocol}. Tables~\ref{tab:families-s2} and~\ref{tab:families-s3} give the gains and oracle gaps by family on the parametric benchmark, at severities above (S2) and below (S3) the calibrated range.

\begin{table}[t]
\centering
\caption{Gain of routing over the best overall denoiser, and oracle gap, by corruption family on the withheld conditions of the discrete benchmark, in dB. Each entry gives gain / oracle gap. The number after each family is its number of withheld conditions; families are weighted by it in the gains of Section~\ref{sec:discrete}. Figure~\ref{fig:families} gives the same gains with 95\% confidence intervals.}
\label{tab:families}
\small
\begin{tabular}{lcccc}
\toprule
Family & Classical & Specialist & Pretrained & Hybrid \\
\midrule
Impulse (1)          & $+6.83$ / 6.88 & $+15.62$ / 15.83 & $+3.91$ / 3.96 & $+13.91$ / 14.07 \\
Banding (2)          & $+6.41$ / 7.95 & $+0.20$ / 2.99   & $+0.55$ / 0.89 & $+6.53$ / 8.19 \\
Correlated (2)       & $+0.76$ / 1.46 & $+1.94$ / 3.81   & $-0.73$ / 0.06 & $-0.72$ / 0.08 \\
Annular (2)          & $+0.71$ / 1.19 & $+1.70$ / 2.15   & $-1.11$ / 0.04 & $-1.13$ / 0.04 \\
Signal-dependent (2) & $+0.23$ / 1.08 & $+1.41$ / 2.11   & $-0.47$ / 0.17 & $-0.52$ / 0.18 \\
Anisotropic (2)      & $+0.11$ / 0.44 & $+0.49$ / 1.81   & $+0.59$ / 1.01 & $+0.58$ / 1.05 \\
Additive (1)         & $-0.76$ / 0.20 & $+0.50$ / 1.33   & $+0.18$ / 0.41 & $+0.14$ / 0.41 \\
Mixture (14)         & $+0.92$ / 1.57 & $+0.96$ / 1.65   & $+0.14$ / 1.10 & $+0.15$ / 1.84 \\
\bottomrule
\end{tabular}
\end{table}

%%%% S2 %%%%%%%

\begin{table}[t]
\centering
\caption{Gain of routing over the best overall denoiser and oracle gap (dB), by corruption family, at severities above the calibrated range (S2) of the parametric benchmark.}
\label{tab:families-s2}
\begin{tabular}{lcccccc}
\toprule
& \multicolumn{2}{c}{Classical}
& \multicolumn{2}{c}{Pretrained}
& \multicolumn{2}{c}{Hybrid} \\
\cmidrule(lr){2-3}
\cmidrule(lr){4-5}
\cmidrule(lr){6-7}
Family & gain & gap & gain & gap & gain & gap \\
\midrule
Gaussian    &  2.39 &  2.59 &  0.14 & 0.31 &   0.10 &  0.31 \\
Laplacian   &  2.18 &  3.15 &  0.37 & 0.53 &   0.30 &  0.53 \\
Uniform     &  2.42 &  2.59 &  0.18 & 0.34 &   0.14 &  0.34 \\
Poisson     &  0.86 &  2.47 & -0.16 & 0.12 &  -0.17 &  0.13 \\
Speckle     & -0.07 &  1.05 & -0.39 & 0.14 &  -0.40 &  0.14 \\
Impulse     & 11.42 & 11.49 &  3.66 & 3.76 &  14.44 & 14.58 \\
Correlated  &  0.38 &  1.20 & -0.90 & 0.04 &  -0.91 &  0.05 \\
Banding     & -0.12 &  0.67 &  0.45 & 0.68 &   0.25 &  0.87 \\
Anisotropic &  0.27 &  0.63 &  0.67 & 0.85 &   0.57 &  0.87 \\
Annular     &  0.57 &  1.71 & -0.91 & 0.10 &  -0.89 &  0.12 \\
\midrule
All ten     &  2.03 &  2.75 &  0.31 & 0.69 &   1.34 &  1.79 \\
\bottomrule
\end{tabular}
\end{table}

%%%% S3 %%%%%%%
\begin{table}[t]
\centering
\caption{Gain of routing over the best overall denoiser and oracle gap (dB), by corruption family, at severities below the calibrated range (S3) of the parametric benchmark.}
\label{tab:families-s3}
\begin{tabular}{lcccccc}
\toprule
& \multicolumn{2}{c}{Classical}
& \multicolumn{2}{c}{Pretrained}
& \multicolumn{2}{c}{Hybrid} \\
\cmidrule(lr){2-3}
\cmidrule(lr){4-5}
\cmidrule(lr){6-7}
Family & gain & gap & gain & gap & gain & gap \\
\midrule
Gaussian    &  0.43 & 2.18 &  0.69 & 1.79 &  0.69 &  1.79 \\
Laplacian   & -0.07 & 1.85 &  0.71 & 1.39 &  0.69 &  1.39 \\
Uniform     &  0.16 & 2.28 &  0.75 & 1.59 &  0.75 &  1.59 \\
Poisson     &  0.13 & 1.16 & -0.01 & 0.41 & -0.01 &  0.41 \\
Speckle     &  2.12 & 4.35 & -0.10 & 1.98 & -0.10 &  1.98 \\
Impulse     &  3.14 & 5.89 &  2.22 & 3.35 &  8.23 & 11.49 \\
Correlated  &  0.45 & 1.09 & -1.04 & 0.02 & -1.04 &  0.03 \\
Banding     &  1.10 & 7.52 &  0.36 & 0.96 &  0.32 &  8.12 \\
Anisotropic & -0.04 & 0.41 &  0.67 & 1.26 &  0.66 &  1.35 \\
Annular     & -0.04 & 1.38 & -0.81 & 0.05 & -0.81 &  0.05 \\
\midrule
All ten     &  0.74 & 2.81 &  0.35 & 1.28 &  0.94 &  2.82 \\
\bottomrule
\end{tabular}
\end{table}

\section{Examples of routing}
\label{app:examples}

\begin{figure}[h]
    \centering
    \includegraphics[width=\linewidth]{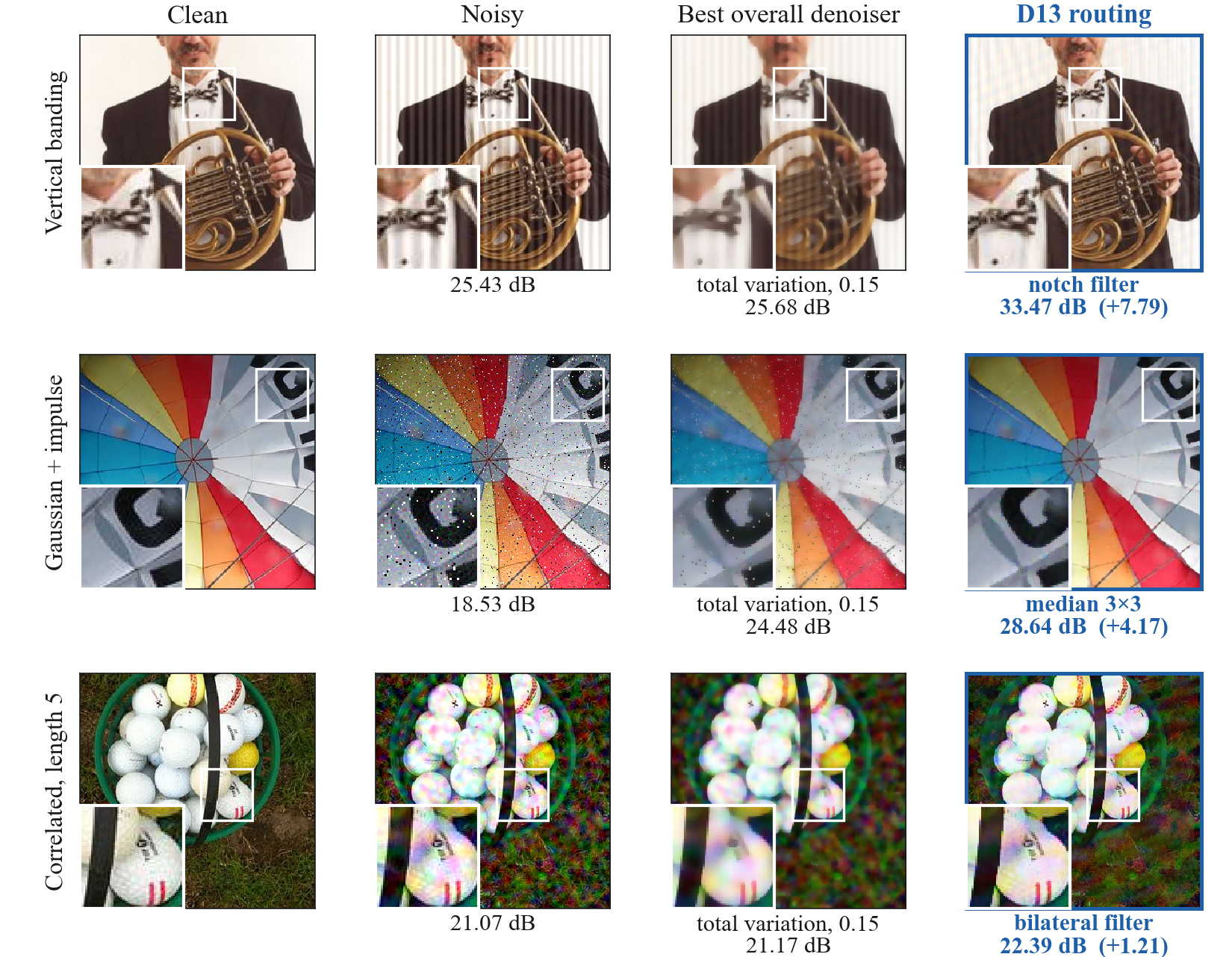}
    \caption{\textbf{Examples of routing.} Each row shows one image and corruption. Columns show the clean image, the noisy image, the output of the best overall denoiser of the classical pool, and the output of the candidate that routing selected. PSNR is given under each panel, and the insets enlarge one region.}
    \label{fig:examples}
\end{figure}

Figure \ref{fig:examples} shows
further withheld conditions with the classical pool, whose best overall denoiser is total
variation. The examples were chosen to illustrate the errors that routing avoids; Section~\ref{sec:families} and Appendix~\ref{app:families}
give the average gains, including the families on which routing gained little or lost PSNR.

\section{CNN Comparison}
\label{app:liu}

We used the network of \citet{liu2020classification} as published, with the SAME padding the paper declares. It has four blocks, each a $3\times3$ convolution, a LeakyReLU with slope 0.1, a second $3\times3$ convolution, a second LeakyReLU, a $2\times2$ max pooling, and batch normalization, at 32, 64, 128, and 256 channels. The $256\times4\times4$ output is flattened and passed through two fully connected layers of 1024 units to four logits. Training used Adam at a learning rate of $10^{-4}$, batches of 128, random flips, and class-balanced epochs, leave-one-content-domain-out over the same ten folds as routing.

The classifier was trained on the eight calibration conditions that fall within its taxonomy, grouped into its four types: Gaussian at $\sigma=15$, 25, and 50; Poisson at peaks 30 and 10; speckle at $\sigma=0.10$ and 0.20; and salt-and-pepper at 5\% density. Routing was restricted to the same four types. Both methods dispatch through the same table of four entries, one denoiser per type, and both run the single candidate the table returns, so they differ only in how the type of an image is chosen.

\paragraph{Conditions compared.} The comparison used the eight withheld conditions composed only of the classifier's four types: four at severities not used in calibration (Gaussian $\sigma=35$, Poisson peak 20, speckle $\sigma=0.15$, salt-and-pepper at 8\%) and four that combine two of them (Gaussian $\sigma=25$ with Poisson peak 30, and Gaussian $\sigma=25$ with salt-and-pepper at 5\%, each at mixing fractions 0.3 and 0.6). The other 18 withheld conditions contain correlated, banding, anisotropic, or annular noise, for which the classifier has no label; they are reported at the end of this appendix.

\paragraph{Preprocessing of the input.} The published method averages the $256\times256$ cell down to $64\times64$ with a $4\times4$ area average. We also report a variant that leaves the amplitude of the noise unchanged: the central $128\times128$ at native resolution, from which training draws random $64\times64$ windows and testing averages the four quadrant tiles in softmax space. On Gaussian noise at $\sigma=25$, the standard deviation of the noise component of the four-phase residual is 25.0 at native resolution and 6.25 after the area average, the factor of four that averaging sixteen samples predicts, and the severities within a family differ only in that amplitude. We report both variants and treat the published one as the faithful reimplementation.

\paragraph{Training data and budget.} We also trained the classifier on nine times the data, so that its accuracy would not be limited by the size of the training set. The $1\times$ arm ran 100 epochs of 26 steps per fold, 2600 steps in all. At nine times the data an epoch is 228 steps, so the arm matched in optimizer steps ran eleven epochs, 2508 steps, 3.5\% short of the $1\times$ budget. A second $9\times$ arm ran 7752 steps, three times the $1\times$ budget. Routing is not trained.

\begin{table}[t]
\centering
\small
\setlength{\tabcolsep}{5pt}
\caption{Gain of routing, restricted to the classifier's four types, over the classifier, on the eight withheld conditions composed only of those types, in dB. Positive favors routing. Bold marks a 95\% interval excluding zero.}
\label{tab:liu-eight}
\begin{tabular}{lrrrr}
\toprule
Arm & Classical & Specialist & Pretrained & Hybrid \\
\midrule
Area average, $1\times$ (published) & $\mathbf{+0.70}$ & $\mathbf{+0.47}$ & $\mathbf{+0.62}$ & $\mathbf{+0.63}$ \\
Area average, $9\times$, 2508 steps & $\mathbf{+0.81}$ & $\mathbf{+0.56}$ & $\mathbf{+0.81}$ & $\mathbf{+0.87}$ \\
Area average, $9\times$, 7752 steps & $\mathbf{+0.84}$ & $\mathbf{+0.66}$ & $\mathbf{+0.88}$ & $\mathbf{+0.99}$ \\
Native resolution, $1\times$        & $\mathbf{+0.33}$ & $-0.01$          & $\mathbf{+0.42}$ & $\mathbf{+0.29}$ \\
Native resolution, $9\times$, 2508 steps & $\mathbf{+0.38}$ & $+0.08$ & $\mathbf{+0.47}$ & $\mathbf{+0.35}$ \\
Native resolution, $9\times$, 7752 steps & $\mathbf{+0.38}$ & $+0.01$ & $\mathbf{+0.44}$ & $\mathbf{+0.33}$ \\
\bottomrule
\end{tabular}
\end{table}

\paragraph{Gain on the eight conditions.} With the published preprocessing, routing gave higher PSNR than the classifier in every pool, by 0.47 to 0.70~dB, with intervals $[0.46, 1.03]$, $[0.26, 0.73]$, $[0.49, 0.75]$, and $[0.46, 0.80]$. Reading the noise at native resolution narrowed the difference without changing its sign in three of the four pools, and in the specialist pool the two drew level, with interval $[-0.13, 0.11]$.

\begin{table}[t]
\centering
\caption{The same comparison split into its two halves: the four conditions that are pure instances of the classifier's types, and the four that combine two of them. Gain of routing over the classifier, in dB; positive favors routing, and bold marks a 95\% interval excluding zero.}
\label{tab:liu-split}

\small
\setlength{\tabcolsep}{2.2pt}
\renewcommand{\arraystretch}{1.02}

\begin{tabular*}{\textwidth}{
    @{\extracolsep{\fill}}
    l
    rrrr
    @{\hspace{1.6em}}
    rrrr
    @{}
}
\toprule
& \multicolumn{4}{c}{Four unseen severities} & \multicolumn{4}{c}{Four mixtures} \\
\cmidrule(lr){2-5}\cmidrule(lr){6-9}
Arm & Clas. & Spec. & Pre. & Hyb. & Clas. & Spec. & Pre. & Hyb. \\
\midrule

Area average, $1\times$ (published)
& $+0.05$ & $-0.04$ & $+0.06$ & $+0.03$
& $\mathbf{+1.35}$ & $\mathbf{+0.99}$ & $\mathbf{+1.17}$ & $\mathbf{+1.22}$ \\

Area average, $9\times$, 2508 steps
& $-0.07$ & $\mathbf{-0.19}$ & $+0.01$ & $-0.05$
& $\mathbf{+1.70}$ & $\mathbf{+1.32}$ & $\mathbf{+1.60}$ & $\mathbf{+1.80}$ \\

Area average, $9\times$, 7752 steps
& $-0.06$ & $-0.19$ & $-0.01$ & $-0.06$
& $\mathbf{+1.75}$ & $\mathbf{+1.51}$ & $\mathbf{+1.77}$ & $\mathbf{+2.04}$ \\

Native resolution, $1\times$
& $\mathbf{-0.20}$ & $\mathbf{-0.31}$ & $-0.02$ & $\mathbf{-0.11}$
& $\mathbf{+0.86}$ & $\mathbf{+0.30}$ & $\mathbf{+0.86}$ & $\mathbf{+0.69}$ \\

Native resolution, $9\times$, 2508 steps
& $\mathbf{-0.17}$ & $\mathbf{-0.28}$ & $-0.01$ & $-0.10$
& $\mathbf{+0.92}$ & $\mathbf{+0.45}$ & $\mathbf{+0.95}$ & $\mathbf{+0.81}$ \\

Native resolution, $9\times$, 7752 steps
& $\mathbf{-0.16}$ & $\mathbf{-0.34}$ & $-0.02$ & $-0.11$
& $\mathbf{+0.92}$ & $\mathbf{+0.37}$ & $\mathbf{+0.89}$ & $\mathbf{+0.77}$ \\

\bottomrule
\end{tabular*}
\end{table}

\paragraph{Where the difference comes from.} On the four conditions that are pure instances of the classifier's types the two methods are close (Table~\ref{tab:liu-split}). With the published preprocessing at the standard budget they do not differ in any pool; nine times the data puts the classifier 0.19~dB ahead in the specialist pool, and at native resolution it leads there by up to 0.34~dB. All of routing's advantage on the eight conditions comes from the four mixtures, where it gains 0.99 to 1.35~dB with the published preprocessing and 0.30 to 0.86~dB at native resolution. The specialist pool separates the two effects: at native resolution with the largest budget the classifier leads by 0.34~dB on the four severities and trails by 0.37~dB on the four mixtures, and the two cancel to the parity recorded in Table~\ref{tab:liu-eight}. A mixture of two of the classifier's types is nameable component by component and has no name as a whole, and the output space has four labels and none for a combination of types, so the classifier must discard one of the two mechanisms present in the image. Routing carries no such restriction, because its references are labeled by the denoiser that gives the highest PSNR rather than by the corruption.

\paragraph{Effect of more training data.} Nine times the training data raised the classifier's type accuracy on unseen severities from 80.4 to 86.9\% at the matched budget (Table~\ref{tab:liu-accuracy}), and it widened routing's advantage on the eight conditions in every pool and under both input variants (Table~\ref{tab:liu-eight}). The improvement in recognition falls on the conditions that have a name in the classifier's taxonomy; the mixtures, which have none, are where the two methods differ, and they are unaffected. Identifying the corruption more accurately therefore did not lead to a better choice of denoiser.

\begin{table}[t]
\centering
\small
\caption{Noise-type accuracy on the four withheld conditions that are pure instances of the classifier's types. A mixture has no single true type, so accuracy is not defined on the other four. With four classes, chance is 0.25.}
\label{tab:liu-accuracy}
\begin{tabular}{lrrrrr}
\toprule
Arm & Gaussian 35 & Poisson 20 & Speckle 0.15 & S\&P 8\% & All four \\
\midrule
Routing, four types                      & $0.967$ & $0.913$ & $0.850$ & $0.950$ & $0.920$ \\
\midrule
Area average, $1\times$ (published)      & $0.683$ & $0.803$ & $0.753$ & $0.977$ & $0.804$ \\
Area average, $9\times$, 2508 steps      & $0.700$ & $0.900$ & $0.883$ & $0.993$ & $0.869$ \\
Area average, $9\times$, 7752 steps      & $0.710$ & $0.863$ & $0.930$ & $0.990$ & $0.873$ \\
Native resolution, $1\times$             & $0.397$ & $0.950$ & $0.970$ & $1.000$ & $0.829$ \\
Native resolution, $9\times$, 2508 steps & $0.457$ & $0.933$ & $0.967$ & $1.000$ & $0.839$ \\
Native resolution, $9\times$, 7752 steps & $0.603$ & $0.983$ & $0.970$ & $0.997$ & $0.888$ \\
\bottomrule
\end{tabular}
\end{table}

\paragraph{Accuracy on interpolated severities.} Gaussian noise at $\sigma=35$ lies between the $\sigma=25$ and $\sigma=50$ the classifier trained on, and it is the condition it identifies least often, the lowest of its four columns in every arm. The published arm names it correctly for 68.3\% of images and the descriptor for 96.7\%. The native-resolution arm, whose accuracy on held-out images of the eight calibration conditions exceeds 99\%, falls to 39.7\% here, so reading the noise at full amplitude let it learn the severities it was shown without letting it interpolate between them. More training raises this column to 45.7\% at the matched budget and to 60.3\% when the optimizer budget is tripled as well, and it remains the lowest of the four throughout.

\paragraph{Images outside the four types.} On the 18 withheld conditions that contain correlated, banding, anisotropic, or annular noise, the classifier must still return one of its four labels. The published arm calls 55.8\% of them Gaussian and 21.4\% Poisson, and splits the remaining 22.8\% almost evenly between speckle and salt-and-pepper. It assigns these labels with high confidence: the mean probability of its top class on these images is 0.929, and 77.7\% of them exceed 0.9. A fixed taxonomy gives no indication that an image falls outside it. The out-of-support score of Appendix~\ref{app:sidd} provides that indication for the descriptor.

\begingroup
\section{Learned alternatives to the descriptor and the router}
\label{app:learned}

Routing has two parts that could be learned instead of designed: the map from a corrupted image to its descriptor, and the rule that maps a descriptor to a candidate. We replaced each with learned alternatives trained on the same data and labels that routing uses, and compared them with \Dxiii and nearest-neighbor retrieval on the 26 withheld conditions of the discrete benchmark. We also compared routing with MoCE-IR \citep{zamfir2025complexity}, a restoration network that selects among experts internally.

\paragraph{Data and protocol.} Every alternative was trained, in each of the ten leave-one-domain-out folds of Section~\ref{sec:protocol}, on the images the nearest-neighbor router uses as references: the 14 calibration conditions in the nine training domains, 30 photographs each, for 3{,}780 images per fold, together with the PSNR of every candidate on each image. The withheld conditions were never used for training or for choosing hyperparameters. Gains and intervals follow Section~\ref{sec:protocol}. Besides the four pools of Section~\ref{sec:pools}, we report a pool of all 36 distinct candidates, whose best overall denoiser is \texttt{S\_SCUNet\_real}.

\paragraph{Learned routers on \Dxiii.} We kept \Dxiii and replaced nearest-neighbor retrieval with four learners. A logistic regression predicts the calibration condition, and the candidate is then read from the same table of best candidates that routing uses. A second logistic regression and a gradient-boosted tree ensemble (200 iterations) predict the best candidate of each training image directly. A ridge regression on quadratic features of \Dxiii predicts the PSNR of every candidate, and the candidate with the highest prediction is run. Hyperparameters were chosen inside each fold on the training domains only, under one of two rules: holding out one training domain, which tests new photographs of known corruptions, or holding out one calibration condition, which tests a corruption not seen in training, as at test time. The second rule was added after a first run that used only the first.

\paragraph{CNN on pixels.} We retrained the network of \citet{liu2020classification}, with the settings of Appendix~\ref{app:liu} and 14 outputs, one per calibration condition, on $128\times128$ crops of the same images, without a hyperparameter search. We read it in two ways. Its softmax selects a condition, and the candidate is read from the table of best candidates. Alternatively, its 1024-dimensional penultimate layer replaces \Dxiii as the space in which nearest-neighbor retrieval is performed.

\begin{table}[h]

\centering
\small
\caption{Routing gain over the best overall denoiser (dB) on the 26 withheld conditions. For each learned alternative, the row gives its best variant in each pool, chosen after the test results were seen, which favors the alternatives. Bold marks a 95\% paired interval of the difference from \Dxiii with nearest-neighbor retrieval that excludes zero.}
\label{tab:learned-summary}
\begin{tabular}{lccccc}
\toprule
Method & Classical & Specialist & Pretrained & Hybrid & All 36 \\
\midrule
\Dxiii, nearest neighbor & $+1.36$ & $+1.58$ & $+0.14$ & $+0.99$ & $+1.20$ \\
Learned router on \Dxiii, best of 8 variants & $+1.36$ & $+1.56$ & $+0.05$ & $+0.85$ & $+1.10$ \\
CNN on pixels, softmax readout & $+1.29$ & $+1.53$ & $\mathbf{-0.11}$ & $+0.97$ & $\mathbf{+1.04}$ \\
\bottomrule
\end{tabular}
\end{table}

\paragraph{Results.} No learned alternative improved on \Dxiii with nearest-neighbor retrieval in any pool (Table~\ref{tab:learned-summary}). All 40 differences between a learned router and nearest-neighbor retrieval were negative, and 31 of them had intervals excluding zero (Table~\ref{tab:learned-routers}). When hyperparameters were chosen by holding out a calibration condition, the learners often selected their strongest regularization, the ridge penalty $\alpha=10^4$ in 30 of 50 folds and the logistic penalty $C=10^{-4}$ in 20 of 50, which makes their choice nearly constant across images. On a corruption absent from training, they found little to gain over a fixed choice. The CNN matched \Dxiii in the classical, specialist, and hybrid pools and fell below it in the pretrained pool and the pool of all 36 candidates. Its embedding was a worse space for retrieval than its softmax (Table~\ref{tab:learned-cnn}), and reducing it to 13 dimensions by principal components, whitening it, or retrieving 5 or 10 neighbors did not help.

\begin{table}[h]

\centering
\caption{Learned routers on \Dxiii: difference in routing gain from nearest-neighbor retrieval (dB), with 95\% paired intervals. The first row gives the gain of nearest-neighbor retrieval over the best overall denoiser. Bold marks an interval that excludes zero. The rule in brackets is the one used to choose hyperparameters.}
\label{tab:learned-routers}
\resizebox{\linewidth}{!}{%
\begin{tabular}{llccccc}
\toprule
Learner & Rule & Classical & Specialist & Pretrained & Hybrid & All 36 \\
\midrule
Nearest neighbor (gain) & & $+1.36$ & $+1.58$ & $+0.14$ & $+0.99$ & $+1.20$ \\
\midrule
Logistic, to condition & [domain] & $\mathbf{-0.22}$ {\scriptsize[$-0.29$, $-0.16$]} & $\mathbf{-0.14}$ {\scriptsize[$-0.25$, $-0.06$]} & $\mathbf{-0.16}$ {\scriptsize[$-0.25$, $-0.08$]} & $\mathbf{-0.32}$ {\scriptsize[$-0.44$, $-0.22$]} & $\mathbf{-0.20}$ {\scriptsize[$-0.32$, $-0.11$]} \\
Logistic, to candidate & [domain] & $\mathbf{-0.23}$ {\scriptsize[$-0.38$, $-0.10$]} & $\mathbf{-0.11}$ {\scriptsize[$-0.19$, $-0.04$]} & $\mathbf{-0.20}$ {\scriptsize[$-0.37$, $-0.05$]} & $\mathbf{-0.27}$ {\scriptsize[$-0.42$, $-0.12$]} & $\mathbf{-0.24}$ {\scriptsize[$-0.32$, $-0.15$]} \\
Boosted trees, to candidate & [domain] & $\mathbf{-0.20}$ {\scriptsize[$-0.39$, $-0.04$]} & $\mathbf{-0.23}$ {\scriptsize[$-0.30$, $-0.15$]} & $\mathbf{-0.21}$ {\scriptsize[$-0.34$, $-0.08$]} & $\mathbf{-0.15}$ {\scriptsize[$-0.27$, $-0.00$]} & $\mathbf{-0.52}$ {\scriptsize[$-0.68$, $-0.37$]} \\
Ridge, PSNR of each candidate & [domain] & $\mathbf{-0.11}$ {\scriptsize[$-0.20$, $-0.02$]} & $\mathbf{-0.10}$ {\scriptsize[$-0.20$, $-0.02$]} & $-0.10$ {\scriptsize[$-0.21$, $0.02$]} & $\mathbf{-0.24}$ {\scriptsize[$-0.37$, $-0.11$]} & $\mathbf{-0.16}$ {\scriptsize[$-0.27$, $-0.03$]} \\
\midrule
Logistic, to condition & [condition] & $\mathbf{-0.12}$ {\scriptsize[$-0.19$, $-0.05$]} & $-0.05$ {\scriptsize[$-0.12$, $0.00$]} & $\mathbf{-0.18}$ {\scriptsize[$-0.29$, $-0.10$]} & $\mathbf{-0.35}$ {\scriptsize[$-0.50$, $-0.25$]} & $\mathbf{-0.13}$ {\scriptsize[$-0.25$, $-0.03$]} \\
Logistic, to candidate & [condition] & $-0.01$ {\scriptsize[$-0.09$, $0.08$]} & $\mathbf{-0.09}$ {\scriptsize[$-0.16$, $-0.03$]} & $-0.14$ {\scriptsize[$-0.27$, $0.00$]} & $\mathbf{-0.98}$ {\scriptsize[$-1.16$, $-0.70$]} & $-0.10$ {\scriptsize[$-0.23$, $0.02$]} \\
Boosted trees, to candidate & [condition] & $\mathbf{-0.28}$ {\scriptsize[$-0.46$, $-0.12$]} & $\mathbf{-0.27}$ {\scriptsize[$-0.38$, $-0.18$]} & $\mathbf{-0.22}$ {\scriptsize[$-0.33$, $-0.11$]} & $-0.14$ {\scriptsize[$-0.27$, $0.01$]} & $\mathbf{-0.50}$ {\scriptsize[$-0.64$, $-0.36$]} \\
Ridge, PSNR of each candidate & [condition] & $-0.08$ {\scriptsize[$-0.16$, $0.00$]} & $-0.02$ {\scriptsize[$-0.12$, $0.08$]} & $\mathbf{-0.15}$ {\scriptsize[$-0.26$, $-0.03$]} & $\mathbf{-0.68}$ {\scriptsize[$-0.83$, $-0.48$]} & $-0.22$ {\scriptsize[$-0.48$, $0.06$]} \\
\bottomrule
\end{tabular}}
\end{table}

\begin{table}[h]

\centering
\caption{CNN on pixels: difference in routing gain from \Dxiii with nearest-neighbor retrieval (dB), with 95\% paired intervals, for the two readouts. Bold marks an interval that excludes zero.}
\label{tab:learned-cnn}
\resizebox{\linewidth}{!}{%
\begin{tabular}{lccccc}
\toprule
Readout & Classical & Specialist & Pretrained & Hybrid & All 36 \\
\midrule
Softmax, then table of best candidates & $-0.07$ {\scriptsize[$-0.14$, $0.01$]} & $-0.05$ {\scriptsize[$-0.16$, $0.07$]} & $\mathbf{-0.25}$ {\scriptsize[$-0.35$, $-0.15$]} & $-0.01$ {\scriptsize[$-0.13$, $0.11$]} & $\mathbf{-0.16}$ {\scriptsize[$-0.29$, $-0.03$]} \\
Nearest neighbor in the embedding & $\mathbf{-0.14}$ {\scriptsize[$-0.24$, $-0.05$]} & $-0.11$ {\scriptsize[$-0.22$, $0.02$]} & $\mathbf{-0.31}$ {\scriptsize[$-0.43$, $-0.19$]} & $-0.06$ {\scriptsize[$-0.18$, $0.11$]} & $\mathbf{-0.27}$ {\scriptsize[$-0.40$, $-0.12$]} \\
\bottomrule
\end{tabular}}
\end{table}

\paragraph{Distance to the references.} The two spaces differ in how far a new corruption lies from the references. In \Dxiii, the median distance from an image to its nearest reference was 0.81 for images of calibration conditions and 1.02 for images of withheld conditions, a ratio of 1.27. In the embedding of the CNN the medians were 3.13 and 8.59, a ratio of 2.74. The network separates the 14 corruptions it was trained on, and images of the withheld conditions fall far from all of them, which is consistent with its embedding being a worse space for retrieval than its softmax.

\paragraph{MoCE-IR.} We used the public all-in-one checkpoint of MoCE-IR, trained on Gaussian noise at $\sigma=15$, 25, and 50, rain, and haze, without modification, after reproducing the published PSNR on CBSD68 at the three noise levels. Applied to every image of the 26 withheld conditions, it reached 24.91~dB. Routing gave higher PSNR in every pool (Table~\ref{tab:moce}), and among all 37 candidates MoCE-IR was the best only on Gaussian noise at $\sigma=35$, the withheld condition closest to its training data. Added as one more candidate to the pretrained and hybrid pools and to the pool of all 36, with no other change to the procedure, it raised the routing gain by 0.06 to 0.08~dB and was selected for 34 to 44\% of images.

\begin{table}[h]

\centering
\small
\caption{MoCE-IR on the 26 withheld conditions. Mean routed PSNR of each pool (dB); difference between routing and MoCE-IR applied to every image, positive when routing is better; change in routing gain when MoCE-IR is added as a candidate; and share of images routed to it. Intervals are 95\% paired intervals.}
\label{tab:moce}
\begin{tabular}{lcccc}
\toprule
 & & & \multicolumn{2}{c}{MoCE-IR as a candidate} \\
\cmidrule(lr){4-5}
Pool & Routed & Routing $-$ MoCE-IR & Change & Selected \\
\midrule
Classical       & 25.73 & $+0.82$ {\scriptsize[0.43, 1.22]} & -- & -- \\
Specialist      & 27.49 & $+2.58$ {\scriptsize[2.43, 2.74]} & -- & -- \\
Pretrained & 26.62 & $+1.71$ {\scriptsize[1.52, 1.92]} & $+0.07$ {\scriptsize[0.03, 0.13]} & 44\% \\
Hybrid          & 27.46 & $+2.55$ {\scriptsize[2.32, 2.77]} & $+0.08$ {\scriptsize[0.04, 0.14]} & 36\% \\
All 36          & 27.68 & $+2.77$ {\scriptsize[2.61, 2.93]} & $+0.06$ {\scriptsize[0.02, 0.11]} & 34\% \\
\bottomrule
\end{tabular}
\end{table}
\endgroup

\begingroup
\section{Ablation of the descriptor}
\label{app:ablation}

To find which properties of the noise determine the choice of denoiser, we removed groups of statistics from \Dxiii and repeated the routing experiment on the 26 withheld conditions of the discrete benchmark, leaving everything else unchanged: the folds, nearest-neighbor retrieval, the table of best candidates, and the best overall denoiser. With all thirteen statistics, the procedure reproduces the gains of Section~\ref{sec:discrete}. The pool of all 36 candidates is the one defined in Appendix~\ref{app:learned}.

\paragraph{Groups.} We used five groups, following the properties of Table~\ref{tab:statistics}: amplitude (\texttt{marg\_var}), marginal law (\texttt{marg\_skew} and \texttt{marg\_kurt}), signal dependence (\texttt{sig\_slope}), spectral structure (\texttt{peak\_prom}, \texttt{line\_frac}, \texttt{ang\_conc}, and \texttt{anisotropy}), and spatial correlation (\texttt{spat\_rlag01} and \texttt{freq\_band0-3}). Table~\ref{tab:statistics} lists the lag-one correlation and the radial bands as separate properties. We merged them into one group after a first run with six groups had been seen, because both describe the second-order statistics of the residual. We also kept only the first three groups, which describe each residual sample separately, or only the last two, which describe spatial structure.

The ablation is diagnostic. Any reduced descriptor that did better than \Dxiii would have been identified on the withheld conditions, so none of the variants replaces it.

\paragraph{Results.} The marginal law contributed most of the gain (Table~\ref{tab:ablation}). Without skewness and kurtosis, routing lost 0.77 to 1.53~dB, depending on the pool, and the loss fell mainly on impulse noise, mixtures, and additive noise (Table~\ref{tab:ablation-family}). Amplitude, spectral structure, and signal dependence each contributed between 0.02 and 0.39~dB on average, and most of these differences had intervals excluding zero. The average understates the role of spectral structure, which is concentrated on two families. Without it, routing lost 0.77 to 2.00~dB on annular noise in every pool, and about 2~dB on banding in the classical and hybrid pools, which contain the notch filter. Annular noise accounts for 2 of the 26 withheld conditions, so a loss of 2~dB on it lowers the average gain by about 0.15~dB. Signal dependence contributed 0.40 to 0.83~dB on the signal-dependent family in every pool except the classical one.

The spatial-correlation group contributed little. Removing it changed the gain by $-0.07$ to $+0.14$~dB, and no family lost more than 0.30~dB. These statistics are computed on a residual that attenuates the low frequencies, where correlated noise places most of its energy (Appendix~\ref{app:sensitivity}), which may explain why they add little to the choice. Their effect also changed sign between pools, and we keep them in \Dxiii.

Neither block was sufficient alone. Keeping only amplitude, marginal law, and signal dependence lost 0.17 to 0.40~dB, and keeping only the spatial statistics lost 0.95 to 1.75~dB, with intervals excluding zero in every pool.

\begin{table}[h]

\centering
\caption{Change in routing gain (dB) on the 26 withheld conditions when a group of statistics is removed, or when only one block is kept, relative to the full descriptor, with 95\% paired intervals. The first row gives the gain of the full descriptor over the best overall denoiser. The number after each variant is the number of statistics removed or kept. Bold marks an interval that excludes zero.}
\label{tab:ablation}
\resizebox{\linewidth}{!}{%
\begin{tabular}{lccccc}
\toprule
Variant & Classical & Specialist & Pretrained & Hybrid & All 36 \\
\midrule
Full \Dxiii (gain) & $+1.36$ & $+1.58$ & $+0.14$ & $+0.99$ & $+1.20$ \\
\midrule
Remove marginal law (2) & $\mathbf{-0.77}$ {\scriptsize[$-0.92$, $-0.65$]} & $\mathbf{-1.11}$ {\scriptsize[$-1.20$, $-1.01$]} & $\mathbf{-0.95}$ {\scriptsize[$-1.05$, $-0.85$]} & $\mathbf{-1.35}$ {\scriptsize[$-1.47$, $-1.23$]} & $\mathbf{-1.53}$ {\scriptsize[$-1.66$, $-1.39$]} \\
Remove amplitude (1) & $-0.06$ {\scriptsize[$-0.13$, $0.01$]} & $\mathbf{-0.39}$ {\scriptsize[$-0.48$, $-0.31$]} & $\mathbf{-0.23}$ {\scriptsize[$-0.34$, $-0.12$]} & $\mathbf{-0.32}$ {\scriptsize[$-0.44$, $-0.21$]} & $\mathbf{-0.32}$ {\scriptsize[$-0.45$, $-0.19$]} \\
Remove spectral structure (4) & $\mathbf{-0.15}$ {\scriptsize[$-0.22$, $-0.07$]} & $\mathbf{-0.13}$ {\scriptsize[$-0.23$, $-0.04$]} & $\mathbf{-0.22}$ {\scriptsize[$-0.33$, $-0.10$]} & $\mathbf{-0.24}$ {\scriptsize[$-0.37$, $-0.06$]} & $\mathbf{-0.18}$ {\scriptsize[$-0.30$, $-0.05$]} \\
Remove signal dependence (1) & $\mathbf{-0.09}$ {\scriptsize[$-0.14$, $-0.04$]} & $\mathbf{-0.17}$ {\scriptsize[$-0.21$, $-0.13$]} & $-0.02$ {\scriptsize[$-0.07$, $0.03$]} & $-0.02$ {\scriptsize[$-0.07$, $0.04$]} & $-0.02$ {\scriptsize[$-0.08$, $0.04$]} \\
Remove spatial correlation (5) & $+0.03$ {\scriptsize[$-0.01$, $0.08$]} & $\mathbf{-0.07}$ {\scriptsize[$-0.13$, $-0.02$]} & $+0.06$ {\scriptsize[$-0.00$, $0.12$]} & $\mathbf{+0.14}$ {\scriptsize[$0.06$, $0.21$]} & $-0.03$ {\scriptsize[$-0.10$, $0.03$]} \\
\midrule
Keep amplitude, marginal law, and signal dependence (4) & $\mathbf{-0.23}$ {\scriptsize[$-0.31$, $-0.14$]} & $\mathbf{-0.40}$ {\scriptsize[$-0.53$, $-0.28$]} & $\mathbf{-0.19}$ {\scriptsize[$-0.27$, $-0.10$]} & $\mathbf{-0.17}$ {\scriptsize[$-0.29$, $-0.02$]} & $\mathbf{-0.35}$ {\scriptsize[$-0.48$, $-0.23$]} \\
Keep spectral structure and spatial correlation (9) & $\mathbf{-0.95}$ {\scriptsize[$-1.07$, $-0.84$]} & $\mathbf{-1.68}$ {\scriptsize[$-1.79$, $-1.57$]} & $\mathbf{-1.16}$ {\scriptsize[$-1.29$, $-1.04$]} & $\mathbf{-1.66}$ {\scriptsize[$-1.80$, $-1.52$]} & $\mathbf{-1.75}$ {\scriptsize[$-1.90$, $-1.60$]} \\
\bottomrule
\end{tabular}}
\end{table}

\begin{table}[h]

\centering
\small
\caption{Change in the routing gain of each corruption family (dB) when a group of statistics is removed, for the families on which each group has its largest effect. Intervals are omitted because some families contain a single withheld condition.}
\label{tab:ablation-family}
\begin{tabular}{llccccc}
\toprule
Group removed & Family & Classical & Specialist & Pretrained & Hybrid & All 36 \\
\midrule
Marginal law        & Impulse          & $-3.26$ & $-7.50$ & $-2.62$ & $-7.88$ & $-9.38$ \\
                    & Mixture          & $-1.03$ & $-1.06$ & $-1.13$ & $-1.32$ & $-1.53$ \\
                    & Additive         & $-0.38$ & $-1.73$ & $-2.38$ & $-3.37$ & $-2.81$ \\
\addlinespace
Spectral structure  & Banding          & $-1.97$ & $+0.13$ & $-0.19$ & $-2.24$ & $+0.13$ \\
                    & Annular          & $-0.77$ & $-1.98$ & $-1.36$ & $-1.30$ & $-2.00$ \\
                    & Anisotropic      & $-0.17$ & $+0.38$ & $-1.02$ & $-1.01$ & $-0.36$ \\
                    & Correlated       & $-0.31$ & $-0.94$ & $-0.47$ & $-0.44$ & $-0.93$ \\
\addlinespace
Signal dependence   & Signal-dependent & $+0.12$ & $-0.40$ & $-0.68$ & $-0.71$ & $-0.83$ \\
\addlinespace
Spatial correlation & Correlated       & $+0.29$ & $+0.83$ & $+0.40$ & $+0.39$ & $+0.92$ \\
                    & Banding          & $+0.98$ & $+0.03$ & $+0.06$ & $+1.11$ & $+0.03$ \\
                    & Annular          & $-0.10$ & $-0.30$ & $+0.05$ & $+0.05$ & $-0.22$ \\
\bottomrule
\end{tabular}
\end{table}
\endgroup

\section{Noise outside calibration}
\label{app:sidd}

\paragraph{Setup.} The queries are the 1{,}280 sRGB validation blocks of SIDD \citep{abdelhamed2018high}, taken from 40 photographs, 32 blocks each. Blocks of one photograph share a scene, a camera, and an ISO setting, so intervals come from a cluster bootstrap over the 40 photographs, 5{,}000 replicates. The real references are 320 patches of the SIDD Small release, from 160 scene instances over 10 scenes and 5 cameras, with ISO values from 50 to 10{,}000, labeled with the best candidate for their ISO band. The two sets are disjoint, and once the patches are present 99.6\% of the blocks retrieve a real reference.

\paragraph{Gain with and without real references.} Table~\ref{tab:sidd} reports both experiments. The first two columns are the gain over the best overall denoiser of the synthetic calibration set, the baseline used in the main text. The last two measure the per-image choice alone, against the candidate with the highest mean PSNR on the added real references, which is what a system would run once those references are available. In the pretrained and hybrid pools that candidate is a network trained on the SIDD training split, disjoint from the validation blocks; the oracle gap above it is small, and routing does not improve on it. The specialist networks were trained on the synthetic families and none had seen real noise. Removing the synthetic conditions from the library, and routing the blocks against the 320 real patches alone, leaves the gain of every pool within its 95\% interval. The synthetic references are almost never the nearest ones once the real patches are present, and they are still required for synthetic queries, on which the mixed library changes the gain by at most 0.04~dB.

\begin{table}[t]
\centering
\small
\setlength{\tabcolsep}{4pt}
\caption{Routing on the SIDD validation blocks, in dB, with 95\% cluster-bootstrap intervals over the 40 photographs.}
\label{tab:sidd}
\begin{tabular}{lrrrr}
\toprule
 & \multicolumn{2}{c}{Over the synthetic baseline} & \multicolumn{2}{c}{Over the real-reference baseline} \\
\cmidrule(lr){2-3}\cmidrule(lr){4-5}
Pool & synthetic refs. & with real refs. & Gain & Oracle gap \\
\midrule
Classical       & $-3.29$ [$-3.78$, $-2.84$] & $-0.15$ & $-0.15$ [$-0.33$, $0.02$]  & 0.31 \\
Specialist      & $-1.01$ [$-1.53$, $-0.51$] & $+0.88$ & $+1.84$ [$1.11$, $2.59$]   & 3.48 \\
Pretrained & $-7.95$ [$-8.87$, $-6.97$] & $+4.84$ & $-0.04$ [$-0.08$, $-0.00$] & 0.09 \\
Hybrid          & $-8.01$ [$-8.92$, $-7.03$] & $+4.78$ & $-0.04$ [$-0.08$, $-0.01$] & 0.23 \\
\bottomrule
\end{tabular}
\end{table}

\paragraph{The out-of-support score.} Let $z$ be the standardized descriptor of a query. We score it by the negative log-likelihood under the best-fitting calibration condition,
\[
s_{\mathrm{out}}(z) = -\max_{c} \log \mathcal{N}(z; \mu_c, \Sigma_c),
\]
where $\mu_c$ and $\Sigma_c$ are the mean and the Ledoit--Wolf shrunk covariance of the references of condition $c$; the shrinkage is needed because each covariance is estimated from a few hundred references in thirteen dimensions. Scores are computed leave-one-domain-out, and the threshold is the 99th percentile of the calibration images, so 1\% of them are flagged by construction. The flag fires on 90.2\% of the SIDD blocks and 7.3\% of the withheld synthetic images, and within the latter on 15.5\% of the single families but 0.2\% of the mixtures, which superpose families already in the calibration set. The distance to the nearest reference, which routing already computes, gives the same ordering but flags less: 54.3\%, 5.1\%, 10.9\%, and 0.1\%.

\paragraph{Routing on flagged images.} Table~\ref{tab:sidd-split} splits the withheld synthetic images at that threshold. The images here are the 7{,}800 withheld synthetic ones, not the SIDD blocks, so the intervals are those of the discrete benchmark, from the paired bootstrap over content domains and photographs of Appendix~\ref{app:benchmarks}; the cluster bootstrap declared
above applies only to Table~\ref{tab:sidd}. The comparison is made in recovered share rather than in decibels, because a flagged image lies farther from the references and its oracle gap is larger; in the specialist pool routing gains more decibels on the flagged images while recovering less of what is available. Splitting on the distance instead gives the same sign in three pools.

\begin{table}[H]
\centering
\small
\setlength{\tabcolsep}{4pt}
\caption{Routing on the withheld synthetic images, split by the flag. \emph{Recovered} is the gain as a share of the oracle gap. The last column gives the difference in recovered share between unflagged and flagged images, in percentage points, with a 95\% interval. It is computed from the unrounded shares, so it does not reproduce the difference of the two rounded columns exactly.}

\label{tab:sidd-split}
\begin{tabular}{lrrrrr}
\toprule
 & \multicolumn{2}{c}{Flagged (566)} & \multicolumn{2}{c}{Unflagged (7234)} & \\
\cmidrule(lr){2-3}\cmidrule(lr){4-5}
Pool & Gain / gap (dB) & Recovered & Gain / gap (dB) & Recovered & Difference \\
\midrule
Classical       & $+1.12$ / 2.38 & 47\%    & $+1.38$ / 2.02 & 68\% & $+22$ [6, 38] \\
Specialist      & $+1.80$ / 3.63 & 50\%    & $+1.56$ / 2.46 & 64\% & $+14$ [2, 27] \\
Pretrained & $-0.32$ / 0.41 & $-78$\% & $+0.18$ / 0.97 & 18\% & $+96$ [36, 180] \\
Hybrid          & $+0.92$ / 2.53 & 36\%    & $+0.99$ / 2.26 & 44\% & $+7$ [$-12$, 36] \\
\bottomrule
\end{tabular}
\end{table}

\begingroup
\section{Corruptions not used to design the descriptor}
\label{app:new-noise}

Appendix~\ref{app:sidd} tests routing on real camera noise. Here we test two further corruptions, neither of which was used to design \Dxiii or appears in the calibration set: Poisson--Gaussian noise, the standard model of a camera sensor, and JPEG compression. Both are applied to the 300 test photographs of the discrete benchmark, with the folds and intervals of Section~\ref{sec:protocol}. Figure~\ref{fig:new-noise} shows where their descriptors fall relative to the benchmark.

\paragraph{Poisson--Gaussian noise.} Each pixel receives Poisson noise with a peak count of 30 or 10 (Appendix~\ref{app:corruptions}) and, on the same pixel, additive Gaussian noise with $\sigma = 5$, 15, or 25, which gives six conditions fixed before the images were generated. This differs from the mixtures of the discrete benchmark, which place Gaussian and Poisson noise on different pixels, so no calibration condition contains both on the same pixel. Nothing was added to the calibration set and nothing was retrained.

The descriptor placed these images next to the Poisson calibration conditions. The nearest reference was a Poisson condition for 87\% of the images, and the Poisson condition with the same peak count for 74\%. The exception was a peak count of 30 with the strongest Gaussian component, which was matched to the stronger Poisson condition, with a peak count of 10, in 49\% of images. The out-of-support score of Appendix~\ref{app:sidd} flagged 0.5\% of the images, fewer than the 1\% of calibration images flagged by construction.

Routing followed this placement (Table~\ref{tab:pg}). In the specialist pool it sent the images to the Poisson networks and gained 3.93~dB over the best overall denoiser, 91\% of the oracle gap. In the pools whose best overall denoiser is \texttt{S\_SCUNet\_real}, the oracle gap was 0.08 to 0.13~dB, because the Poisson conditions are themselves labeled with that network, and routing lost 0.20 to 0.27~dB on the images matched to other references. In the classical pool the oracle gap was 0.28~dB and routing lost 0.06~dB.

\begin{table}[h]
\centering
\small
\caption{Poisson--Gaussian noise, six conditions together. Mean PSNR (dB) of the best overall denoiser of each pool, of routing, and of the oracle, and the routing gain with its 95\% paired interval. Nothing was added to the calibration set.}
\label{tab:pg}
\begin{tabular}{llcccc}
\toprule
Pool & Best overall & Fixed & Routed & Oracle & Gain \\
\midrule
Classical       & \texttt{tv\_0.15}        & 23.68 & 23.63 & 23.96 & $\mathbf{-0.06}$ {\scriptsize[$-0.13$, $-0.01$]} \\
Specialist      & \texttt{M\_laplacian\_25} & 23.20 & 27.13 & 27.52 & $\mathbf{+3.93}$ {\scriptsize[$3.41$, $4.61$]} \\
Pretrained & \texttt{S\_SCUNet\_real} & 28.50 & 28.30 & 28.58 & $\mathbf{-0.20}$ {\scriptsize[$-0.43$, $-0.06$]} \\
Hybrid          & \texttt{S\_SCUNet\_real} & 28.50 & 28.27 & 28.58 & $\mathbf{-0.23}$ {\scriptsize[$-0.55$, $-0.06$]} \\
All 36          & \texttt{S\_SCUNet\_real} & 28.50 & 28.23 & 28.62 & $\mathbf{-0.27}$ {\scriptsize[$-0.65$, $-0.06$]} \\
\bottomrule
\end{tabular}
\end{table}

\paragraph{JPEG compression.} We compressed the 300 test photographs ($256\times256$ crops) with PIL at qualities 20, 10, and 5, with 4:2:0 chroma subsampling. The descriptor flagged 45.2\% of the JPEG images, against 7.3\% of the images of withheld synthetic conditions, and the share increased with compression, from 33.7\% at quality 20 to 42.0\% at quality 10 and 60.0\% at quality 5. 

Without JPEG references, routing sent the JPEG images to candidates chosen for synthetic noise and fell below the best overall denoiser (Table~\ref{tab:jpeg}). We then extended the system in the way Section~\ref{sec:sidd} describes, without retraining any candidate. We added as references the JPEG versions of the calibration photographs of the nine training domains of each fold, 270 images per quality, with one label per quality mapped to its best candidate. We also added one candidate designed for compression artifacts, the public color model of FBCNN \citep{jiang2021towards}, used as released; it estimates the quality itself, so it needs no information about the image. FBCNN became the 15th candidate of the specialist pool and the 37th of the pool of all candidates, and every JPEG label mapped to it in every fold. A JPEG network that we trained ourselves stayed below \texttt{S\_SCUNet\_real} and was not used.

With the references and FBCNN, routing gained 1.51~dB over the best overall denoiser in the specialist pool and 0.52~dB in the pool of all 37 candidates, recovering 86 and 67\% of the oracle gap, and sent 88\% of the JPEG images to FBCNN. When the references of the tested quality were left out, the gain was 1.44 and 0.46~dB. The added references lowered the gain on the 26 withheld synthetic conditions by 0.28 and 0.26~dB, because 7.8\% of those images, mostly from correlated-noise and banding conditions, had a JPEG reference as their nearest neighbor and were sent to FBCNN.

\begin{table}[t]
\centering
\small
\caption{JPEG compression at qualities 20, 10, and 5 on the 300 test photographs. Routing gain over the best overall denoiser of each pool (dB), with 95\% paired intervals. The last row is the change in routing gain on the 26 withheld synthetic conditions after the JPEG references are added. Bold marks an interval that excludes zero.}
\label{tab:jpeg}
\begin{tabular}{lcc}
\toprule
 & Specialist + FBCNN (15) & All + FBCNN (37) \\
\midrule
No JPEG references & $\mathbf{-1.06}$ {\scriptsize[$-1.28$, $-0.89$]} & $\mathbf{-2.04}$ {\scriptsize[$-2.33$, $-1.81$]} \\
JPEG references & $\mathbf{+1.51}$ {\scriptsize[$1.28$, $1.77$]} & $\mathbf{+0.52}$ {\scriptsize[$0.36$, $0.68$]} \\
JPEG references, tested quality left out & $\mathbf{+1.44}$ {\scriptsize[$1.20$, $1.72$]} & $\mathbf{+0.46}$ {\scriptsize[$0.28$, $0.64$]} \\
Share of the oracle gap recovered & 86\% & 67\% \\
JPEG images routed to FBCNN & 88\% & 88\% \\
\midrule
Change on the 26 withheld conditions & $\mathbf{-0.28}$ {\scriptsize[$-0.34$, $-0.22$]} & $\mathbf{-0.26}$ {\scriptsize[$-0.33$, $-0.20$]} \\
\bottomrule
\end{tabular}
\end{table}

\begin{figure}[h]
\centering
{\color{black}\includegraphics[width=\linewidth]{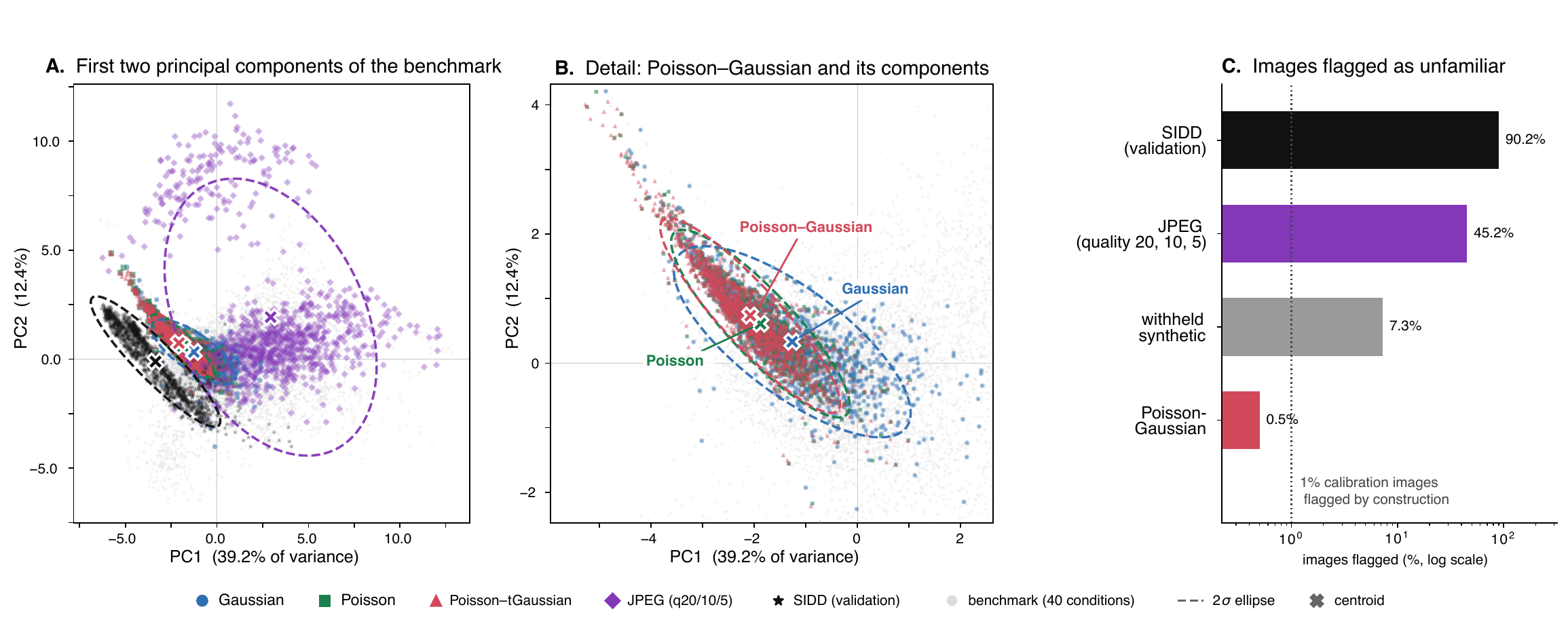}}
\caption{\textbf{Corruptions not used to design \Dxiii.} (A)~Descriptors of the four groups projected onto the first two principal components of the discrete benchmark, which account for 39\% and 12\% of the variance. Dashed lines are $2\sigma$ ellipses and crosses mark centroids. (B)~Detail of the Poisson--Gaussian images and the Gaussian and Poisson calibration conditions. The Poisson--Gaussian centroid lies 0.18 from the Poisson centroid. (C)~Share of images flagged as unfamiliar by the out-of-support score (log scale). The dotted line marks the 1\% of calibration images flagged by construction. The score uses all thirteen dimensions, so images that overlap the benchmark in this projection can still be flagged.}
\label{fig:new-noise}
\end{figure}
\endgroup

\end{document}